\documentclass[10pt,twocolumn,letterpaper]{article}

\usepackage[final]{cvpr}              
\usepackage{multirow}

\definecolor{cvprblue}{rgb}{0.21,0.49,0.74}
\usepackage[pagebackref,breaklinks,colorlinks,allcolors=cvprblue]{hyperref}
\usepackage{booktabs}
\usepackage{pifont}

\def\paperID{} 
\def\confName{CVPR}
\def\confYear{2026}

\title{One Model for All: Unified Try-On and Try-Off in Any Pose via LLM-Inspired Bidirectional Tweedie Diffusion}

\author{
  Jinxi Liu$^{1,\dag}$ \quad
  Zijian He$^{1,\dag}$ \quad
  Guangrun Wang$^{1,2,3,*}$ \quad
  Guanbin Li$^{1,2}$ \quad
  Liang Lin$^{1,2,3}$\\
  $^1$Sun Yat-sen University \quad
  $^2$Guangdong Key Laboratory of Big Data Analysis and Processing \\
  $^3$X-Era AI Lab\\
  {\tt\small \{liujx233, hezj39\}@mail2.sysu.edu.cn,} \\ {\tt\small liguanbin@mail.sysu.edu.cn, linliang@ieee.org, wanggrun@gmail.com}
}

\begin{document}
\twocolumn[{%
\renewcommand\twocolumn[1][]{#1}%
\maketitle
\begin{center}
    \centering
    \captionsetup{type=figure}
    \includegraphics[width=0.88\textwidth]{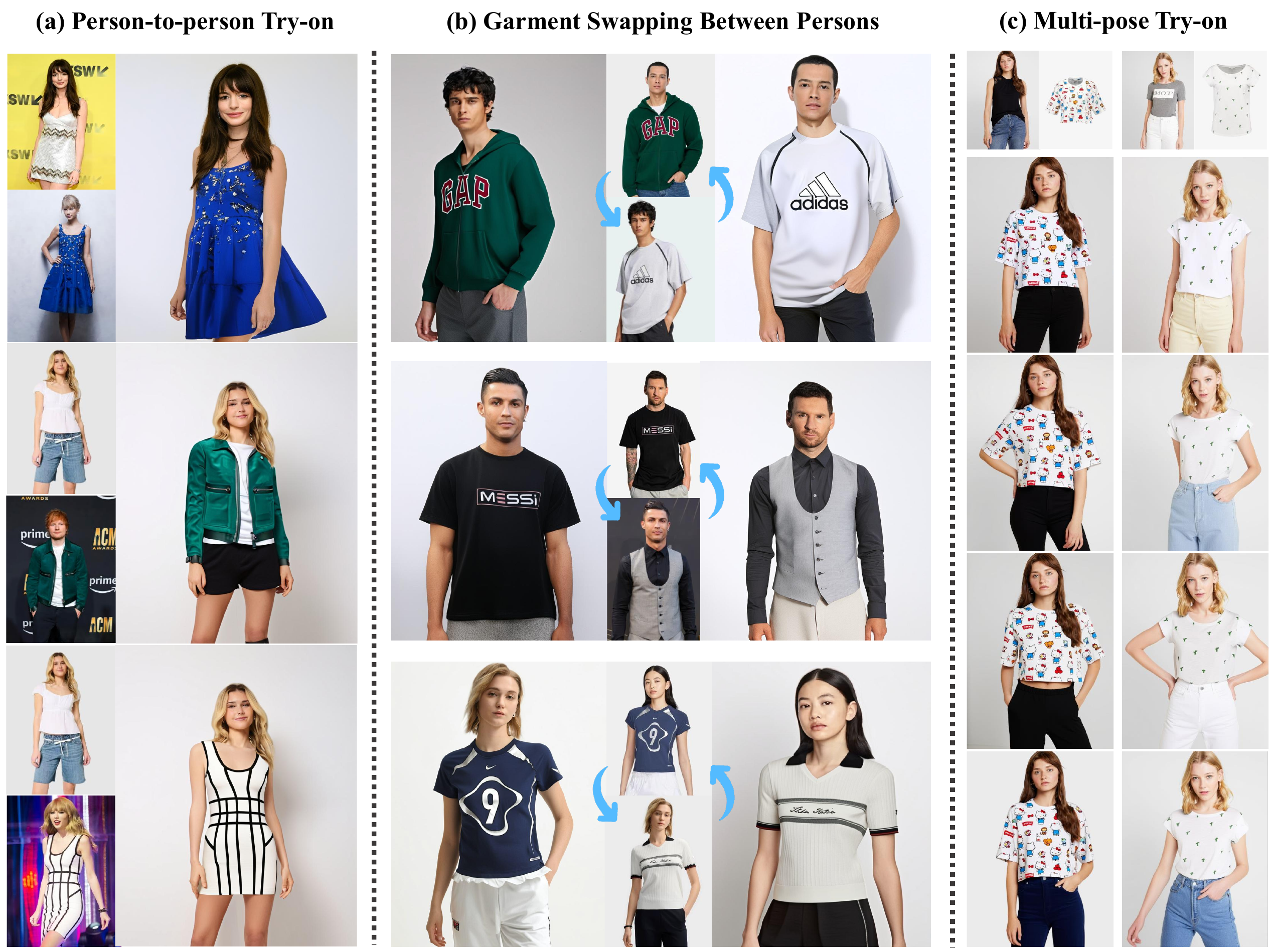}
    \vspace{-5pt}
    \captionof{figure}{
     \textbf{Outfitted models generated by OMFA.} (a) Person-to-person try-on, (b) garment swapping between persons and (c) multi-pose try-on. \emph{Please zoom in to better observe the details.}
     }
    \vspace{-5pt}
    \label{fig:Intro}
\end{center}
}]

\begingroup
  \renewcommand\thefootnote{\fnsymbol{footnote}} 
  \footnotetext[1]{Corresponding author: Guangrun Wang.}
  \footnotetext[2]{These two authors contributed equally and share first authorship.}
\endgroup

\begin{abstract}
Recent diffusion-based approaches have made significant advances in image-based virtual try-on, enabling more realistic and end-to-end garment synthesis. However, most existing methods remain constrained by their reliance on exhibition garments and segmentation masks, as well as their limited ability to handle flexible pose variations. These limitations reduce their practicality in real-world scenarios—for instance, users cannot easily transfer garments worn by one person onto another, and the generated try-on results are typically restricted to the same pose as the reference image. In this paper, we introduce \textbf{OMFA} (\emph{One Model For All}), a unified diffusion framework for both virtual try-on and try-off that operates without the need for exhibition garments and supports arbitrary poses.
OMFA is inspired by the mask-based paradigm of discrete diffusion language models and unifies try-on and try-off within a bidirectional framework. It is built upon a Bidirectional Tweedie Diffusion process for target-selective denoising in latent space.
Instead of imposing lower body constraints, OMFA is an entirely mask-free framework that requires only a single portrait and a target garment as inputs, and is designed to support flexible outfit combinations and cross-person garment transfer, making it better aligned with practical usage scenarios. 
Additionally, by leveraging SMPL-X–based pose conditioning, OMFA supports multi-view and arbitrary-pose try-on from just one image. Extensive experiments demonstrate that OMFA achieves state-of-the-art results on both try-on and try-off tasks, providing a practical and generalizable solution for virtual garment synthesis. Project page: \url{https://onemodelforall.github.io}
\end{abstract}    
\section{Introduction}
\label{sec:intro}

Virtual Try-On (VTON) is the task of generating photorealistic images of a person wearing a desired garment. While recent diffusion-based approaches~\cite{rombach2022high,zhang2023adding,ye2023ip,chen2024anydoor} have significantly advanced this field by improving image realism and enabling end-to-end generation, their applicability remains limited. In particular, most existing methods~\cite{choi2024improving,xu2025ootdiffusion,lee2022hrviton,kim2024stableviton, gou2023taming} rely on clean garment templates (also referred to as exhibition garments) and segmentation masks to isolate clothing regions. These requirements are often impractical in real-world scenarios, such as mobile shopping platforms or social media, where users typically provide casual images without clean garment inputs or multiple views. Moreover, current methods are generally restricted to the pose of the input image, limiting users’ ability to visualize try-on results in arbitrary or user-specified poses.

To alleviate the impractical need for exhibition garments, some methods~\cite{he2024wildvidfit,cui2024street} attempt to extract garments directly from images using segmentation algorithms. However, these techniques often suffer from boundary artifacts, occlusion issues, and garment shape distortion, which degrade the final image quality. Alternatively, several recent works~\cite{velioglu2024try,xarchakos2024tryoffanyone} reformulate try-on as a \emph{try-off} task, aiming to generate a canonical garment representation from a dressed person image. Yet, these approaches treat try-on and try-off as independent tasks, overlooking their inherent duality and failing to provide a unified solution. To address the limitation of pose rigidity, 3D virtual try-on methods~\cite{he2025vton360,xie2024dreamvton} have been explored using parametric body models. However, due to the limited availability of high-quality 3D data compared to 2D imagery, these methods still struggle to produce realistic and high-resolution results.

To overcome these limitations, we propose \textbf{OMFA} (\emph{One Model For All}), a unified diffusion-based framework for both virtual try-on and try-off. Unlike previous methods, OMFA models garment-person transformations bidirectionally within a single architecture, eliminating the need for garment templates or segmentation masks. 
Inspired by the mask-based paradigm of discrete diffusion language models \cite{sahoo2024simple,nie2025large}, 
OMFA represents each modality as latent tokens and unifies try-on and try-off as a latent completion problem. At its core, a \emph{Bidirectional Tweedie Diffusion} process realizes this formulation in continuous latent space by diffusing only the selected target latent while retaining the remaining latents as observed conditions.A shared denoising network is used to handle both generation directions under the same framework, so that the two tasks differ only in the choice of the target latent. 
Furthermore, by incorporating SMPL-X–based~\cite{pavlakos2019expressive} pose conditioning, OMFA enables multi-view and arbitrary-pose try-on from a single image while preserving identity consistency and garment fidelity.

OMFA operates in a practical mask-free setting, requiring only a single portrait image and a target garment image at inference time. Here, ``mask-free'' means that no segmentation or parsing masks are required during inference. This design is motivated by two practical considerations: the target garment may not always be compatible with the original outfit, and users often prefer flexible combinations of upper and lower garments. Therefore, OMFA does not enforce strict preservation of the non-edited garment or the original pose, allowing more flexible and realistic synthesis.

\paragraph{Contributions.} The key contributions of this work are as follows:
\begin{itemize}
\item We introduce \textbf{OMFA}, a unified framework that jointly performs both virtual try-on and try-off within a single architecture, enabling bidirectional garment editing without reliance on segmentation masks or template garments.
\item We formulate try-on and try-off as target-selective latent completion under a shared conditional diffusion framework, and instantiate this unified formulation with Bidirectional Tweedie Diffusion.
\item We incorporate SMPL-X–based pose conditioning to enable arbitrary-pose and multi-view try-on generation from a single portrait image, enhancing the realism and controllability of try-on synthesis.
\item We conduct comprehensive experiments on VITON-HD and Deepfashion-Multimodal datasets, achieving state-of-the-art performance across try-on and try-off in both qualitative and quantitative evaluations. 
\end{itemize}

\section{Related Work}
\label{sec:related}

\paragraph{Image Virtual Try-On.}
Traditional image-based virtual try-on methods mainly rely on GANs with a two-stage pipeline of garment warping and blending. Although techniques such as TPS \cite{yang2020towards, wang2018toward}, optical flow \cite{lee2022hrviton, xie2023gp}, and human parsing priors \cite{choi2021viton, ge2021parser, issenhuth2020not} improve alignment, these methods remain sensitive to pose variation and garment complexity, often producing fusion artifacts and suffering from limited generalization.

With the success of diffusion models in image synthesis, recent virtual try-on research has increasingly shifted toward diffusion-based frameworks. Early methods such as LaDI-VTON \cite{morelli2023ladi} and DCI-VTON \cite{gou2023taming} still rely on explicit garment warping, while later approaches, including TryOnDiffusion \cite{zhu2023tryondiffusion} and subsequent works \cite{kim2024stableviton, xu2025ootdiffusion, choi2024improving}, move toward one-stage generation with improved conditioning mechanisms for implicit alignment and blending. Other methods further explore efficiency \cite{chong2024catvton}, multi-view synthesis \cite{wang2025mv}, and Diffusion Transformer architectures \cite{jiang2024fitdit}. Despite these advances, most diffusion-based try-on models still depend on clean garment templates or segmentation masks, which are impractical in real-world scenarios, and their results remain constrained by the input pose, limiting arbitrary-pose or multi-view visualization.

\vspace{-11pt}
\paragraph{Image Virtual Try-Off.}
To address the dependence on clean garment templates, several recent works \cite{velioglu2024try,xarchakos2024tryoffanyone} have explored the task of image-based virtual try-off—i.e., recovering a canonical garment image from a person wearing the clothing. These methods aim to bypass the need for separate garment inputs by extracting garments directly from dressed images. However, most try-off approaches are limited to generating static garment representations and do not integrate naturally with try-on synthesis. They often treat try-on and try-off as separate pipelines, missing the opportunity to unify both tasks into a single generative model. Additionally, try-off models rarely address pose variation or the generation of garments on new identities, limiting their adaptability in dynamic or user-controllable settings.

\vspace{-11pt}
\paragraph{3D Virtual Try-On.}
To improve pose flexibility, 3D-based virtual try-on approaches explicitly \cite{bridson2002robust,pons2017clothcap} or implicitly model human geometry \cite{he2025vton360}. These methods facilitate pose control and multi-view synthesis. However, due to the scarcity of high-quality 3D garment datasets, these models often suffer from low realism in garment texture and shape fidelity, partly due to the sim-to-real gap. Moreover, 3D fitting pipelines are computationally expensive and less suited for real-time or consumer-facing applications.

\vspace{-11pt}
\paragraph{Summary.}
In summary, current methods either depend on hard-to-obtain garment inputs, lack flexibility in pose, or treat try-on and try-off as separate problems. In contrast, our proposed OMFA addresses all these limitations by unifying try-on and try-off tasks under a single diffusion-based architecture. OMFA eliminates the need for garment templates and segmentation masks, and enables controllable, pose-aware generation via a bidirectional Tweedie diffusion mechanism and SMPL-X conditioning, making it well-suited for practical deployment.

\begin{figure*} [t]
\centering
\includegraphics[width=.9\linewidth]{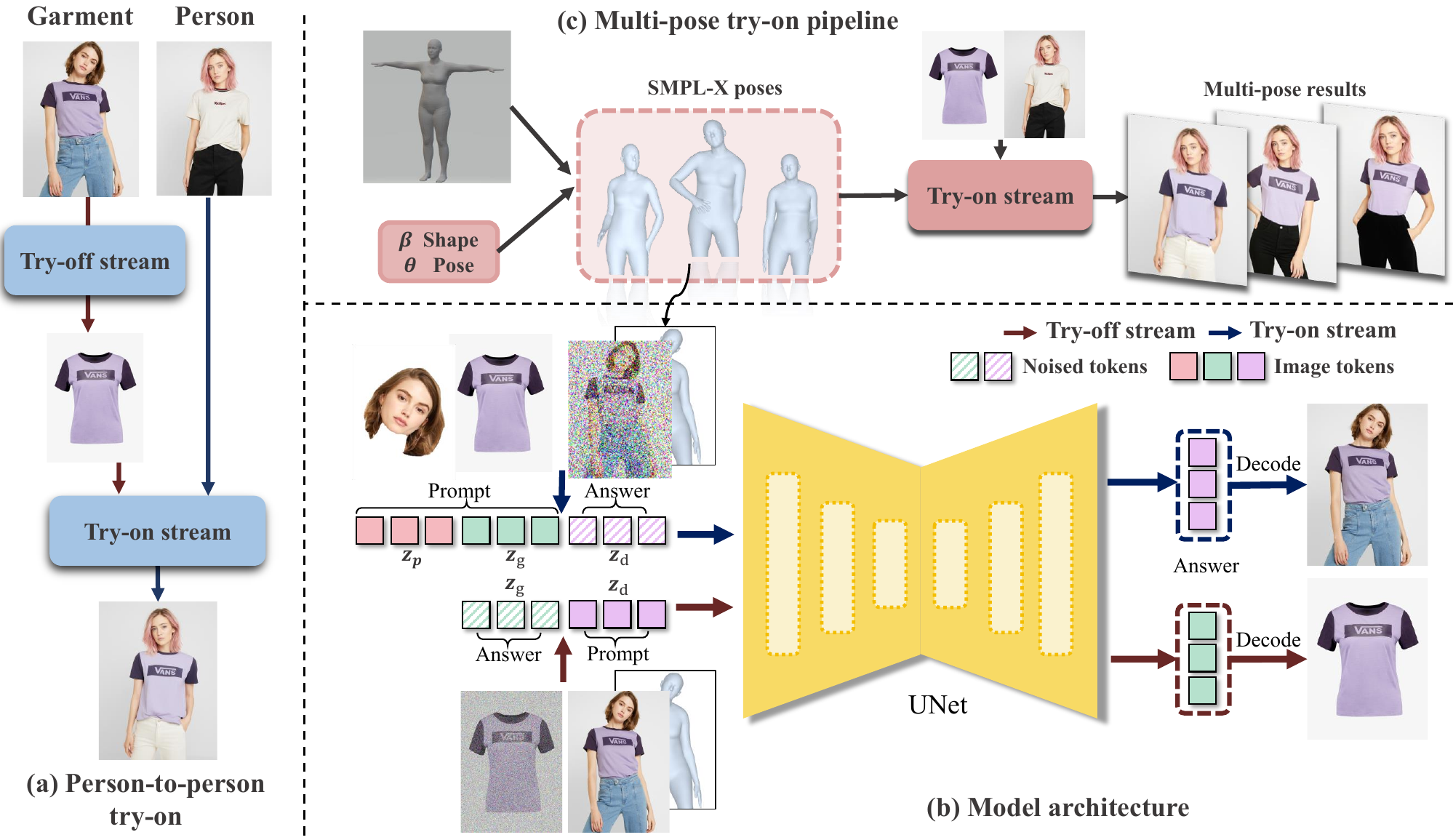} 
    \caption{\textbf{Overview of our proposed OMFA (One Model For All) framework.} (a) illustrates the pipeline of person-to-person try-on, including two processes of try-off and try-on in one model. (b) depicts a model design based on the Tweedie bidirectional diffusion. The model's inputs are the latent token sequence, with noise added to the dressed-person image (try-on stream) or the garment image (try-off stream), while the remaining latents are retained as observed conditions. (c) presents the multi-pose try-on support of our framework.}
    \vspace{-10pt}
 \label{fig:pipeline}
\end{figure*}

\section{Method}
\label{sec:method}

\subsection{Overview}
We present \textbf{OMFA (One Model For All)}, a unified diffusion framework that jointly addresses both virtual \textit{try-on} and \textit{try-off} within a single model.
Given a person image $I_p$, a garment image $I_g$, and a dressed person image $I_d$, OMFA maps them into a shared latent space, where their semantic dependencies can be modeled under a common conditional generation framework.
Within this shared space, try-on and try-off are treated as two task instances of the same latent generation problem: one latent subset is selected as the prediction target, while the remaining subsets are retained as observed conditions.

To instantiate this idea, we formulate generation as conditional latent completion from a masked diffusion perspective, where a selected target latent is recovered from the remaining observed latents (Sec.~\ref{sec:cond_gen}).
Building on this formulation, we further introduce \emph{Bidirectional Tweedie Diffusion} in continuous latent space, which provides a unified theoretical interpretation of target-selective denoising for both generation directions (Sec.~\ref{sec:tweedie}).
Finally, we incorporate explicit structural guidance through an SMPL-X-based conditioning mechanism, which injects 3D human geometry into the latent generation process and enables pose-controllable and shape-consistent synthesis (Sec.~\ref{sec:smpl}).

\subsection{Conditional Generation as Diffusion Language Modeling}
\label{sec:cond_gen}

Let $\mathcal{E}(\cdot)$ and $\mathcal{D}(\cdot)$ denote the encoder and decoder of a latent diffusion autoencoder (e.g., a VAE). We encode the person image $I_p$, garment image $I_g$, and dressed image $I_d$ into latent variables:
\begin{equation}
    \mathbf{z}_p = \mathcal{E}(I_p), \quad \mathbf{z}_g = \mathcal{E}(I_g), \quad \mathbf{z}_d = \mathcal{E}(I_d).
\end{equation}
Rather than treating image domains as fixed inputs and outputs, we formulate these variables as a sequence of generation elements: $\mathcal{Z} = (\mathbf{z}_p, \mathbf{z}_g, \mathbf{z}_d)$. Our method is inspired by diffusion language modeling \cite{nie2025large,sahoo2024simple,xu2026bridgingdiscretecontinuousgapunified}, which excels at generating missing tokens at arbitrary positions within a sequence. Let $\mathcal{Z}^{\mathrm{mask}}$ denote a subset of missing or unobserved latents in the sequence, and $\mathcal{Z}^{\mathrm{obs}} = \mathcal{Z} \setminus \mathcal{Z}^{\mathrm{mask}}$ denote the remaining observed context. Generation is cast as a sequence completion task, where a shared conditional network $f_\theta(\cdot)$ learns a unified rule to recover the missing subset:
\begin{equation}
    p_\theta(\mathcal{Z}^{\mathrm{mask}} \mid \mathcal{Z}^{\mathrm{obs}}) = f_\theta(\mathcal{Z}^{\mathrm{obs}}). \label{eq:mask_gen}
\end{equation}
This sequence-masking formulation naturally unifies virtual try-on and try-off, as they simply correspond to different missing positions in the sequence. For virtual try-on, the missing element is the dressed-person latent: $\mathcal{Z}^{\mathrm{mask}} = \{\mathbf{z}_d\}$, conditioned on $\mathcal{Z}^{\mathrm{obs}} = \{\mathbf{z}_p, \mathbf{z}_g\}$. Conversely, for virtual try-off, the missing element is the garment latent: $\mathcal{Z}^{\mathrm{mask}} = \{\mathbf{z}_g\}$, conditioned on $\mathcal{Z}^{\mathrm{obs}} = \{\mathbf{z}_d\}$. Therefore, OMFA functions as an arbitrary-position latent completion model. However, unlike traditional discrete diffusion language models which do not follow Tweedie's formula, we map this sequence-completion concept into a continuous latent space.

\subsection{Bidirectional Tweedie Diffusion}
\label{sec:tweedie}

Following Sec.~\ref{sec:cond_gen}, we formulate OMFA as a \emph{Tweedie Diffusion Language Model}. Let $m \in \{d, g\}$ denote the selected missing target type in the sequence, where $\mathbf{z}_0^m$ denotes the clean target latent ($\mathcal{Z}^{\mathrm{mask}}$), and $\mathcal{C}^m$ denotes the remaining clean variables used as conditions ($\mathcal{Z}^{\mathrm{obs}}$). We instantiate this masked modeling process by applying continuous Gaussian corruption exclusively to the selected target latent at the missing position, keeping the conditioning variables clean:\begin{equation}q(\mathbf{z}_t^m \mid \mathbf{z}_0^m) = \mathcal{N}!\left( \mathbf{z}_t^m; \sqrt{\bar{\alpha}_t}\mathbf{z}_0^m,, (1-\bar{\alpha}_t)\mathbf{I} \right), \label{eq:forward_target}\end{equation}where $t$ denotes the diffusion timestep and $\bar{\alpha}_t$ denotes the cumulative signal coefficient. Equivalently, the noisy target latent is reparameterized as:\begin{equation}\mathbf{z}_t^m = \sqrt{\bar{\alpha}_t}\mathbf{z}_0^m + \sqrt{1-\bar{\alpha}_t},\epsilon, \qquad \epsilon \sim \mathcal{N}(\mathbf{0}, \mathbf{I}). \label{eq:reparam_target}\end{equation}The fundamental advantage of performing this sequence completion in a continuous domain is that it allows us to mathematically recover the exact posterior mean of the missing content via the \textbf{Tweedie identity} \cite{li2025insitutweediediscretediffusion,zhan2026e0enhancinggeneralizationfinegrained}. Under the conditional noisy latent distribution $p_t(\mathbf{z}_t^m \mid \mathcal{C}^m)$, Tweedie's formula dictates that the expected value of the clean target is strictly defined by the score of the noisy distribution, $s_t^m(\mathbf{z}_t^m,\mathcal{C}^m) := \nabla_{\mathbf{z}_t^m}\log p_t(\mathbf{z}_t^m \mid \mathcal{C}^m)$:\begin{equation}
\begin{aligned}
    &\mathbb{E}[\mathbf{z}_0^m \mid \mathbf{z}_t^m, \mathcal{C}^m] \\ = &\frac{1}{\sqrt{\bar{\alpha}_t}} \left( \mathbf{z}_t^m + (1-\bar{\alpha}t)\nabla{\mathbf{z}_t^m}\log p_t(\mathbf{z}_t^m \mid \mathcal{C}^m) \right).
\end{aligned} \label{eq:tweedie_score}\end{equation}To compute this score, we rely on Denoising Score Matching \cite{vincent2011connection}, which establishes a rigid identity connecting the continuous score field to the expected injected noise:\begin{equation}\nabla_{\mathbf{z}_t^m}\log p_t(\mathbf{z}_t^m \mid \mathcal{C}^m) = -\frac{1}{\sqrt{1-\bar{\alpha}_t}} \mathbb{E}[\epsilon \mid \mathbf{z}_t^m, \mathcal{C}^m]. \label{eq:score_identity}\end{equation}Substituting Eq.~\eqref{eq:score_identity} directly into Tweedie's formula (Eq.~\eqref{eq:tweedie_score}) yields the exact closed-form identity for recovering the clean target latent entirely in terms of the expected noise:\begin{equation}\mathbb{E}[\mathbf{z}_0^m \mid \mathbf{z}_t^m, \mathcal{C}^m] = \frac{1}{\sqrt{\bar{\alpha}_t}} \left( \mathbf{z}_t^m - \sqrt{1-\bar{\alpha}_t},\mathbb{E}[\epsilon \mid \mathbf{z}_t^m, \mathcal{C}^m] \right). \label{eq:tweedie_expected_noise}\end{equation}This theoretical formulation directly dictates our practical implementation. To evaluate Eq.~\eqref{eq:tweedie_expected_noise}, we must estimate the conditional expectation of the noise, $\mathbb{E}[\epsilon \mid \mathbf{z}_t^m, \mathcal{C}^m]$. By standard statistical decision theory, the optimal parameterized function $\epsilon_\theta^*$ that exactly matches a conditional expectation is obtained by minimizing the Mean Squared Error (MSE). Therefore, we implement our shared denoising network $\epsilon_\theta$ using the standard MSE noise-prediction objective:\begin{equation}\mathcal{L}_{\mathrm{diff}} = \mathbb{E}_{m,t,\epsilon} \left[ \left| \epsilon - \epsilon_\theta(\mathbf{z}_t^m, t, \mathcal{C}^m) \right|_2^2 \right]. \label{eq:diff_loss}\end{equation}By minimizing Eq.~\eqref{eq:diff_loss}, the network intrinsically converges to the conditional expectation required by Eq.~\eqref{eq:tweedie_expected_noise}. Thus, optimizing the standard noise prediction loss is not an arbitrary choice, but the mathematically exact implementation required to evaluate Tweedie's sequence-recovery formula. At inference, the target latent is iteratively denoised under the fixed condition set $\mathcal{C}^m$ using the network $\epsilon_\theta$. The final recovered latent is then decoded to image space:\begin{equation}\hat{I}_d = \mathcal{D}(\hat{\mathbf{z}}_0^d), \qquad \hat{I}_g = \mathcal{D}(\hat{\mathbf{z}}_0^g). \label{eq:decode}\end{equation}

\subsection{SMPL-X-based Structural Conditioning}
\label{sec:smpl}
To introduce explicit and controllable human geometric information into the generation process, we propose a SMPL-X-based structural conditioning mechanism, where the rendered structural image $I_s$ serves as the geometric condition. SMPL-X is a low-dimensional parametric human body model that jointly uses shape parameters $\beta \in \mathbb{R}^{10}$ and pose parameters $\theta \in \mathbb{R}^{24 \times 3 \times 3}$ to produce a 3D human mesh with $N=6890$ vertices. Given a person image, we use 4D-Humans \cite{goel2023humans} to regress the corresponding shape parameters $\beta$, pose parameters $\theta$, and camera parameters $\pi$. The estimated SMPL-X mesh is then rendered into an RGB structural image $I_s$ through the camera projection function $\Pi$, formulated as
\begin{equation}
I_s = \Pi(\text{SMPL-X}(\beta, \theta), \pi).
\end{equation}
The rendered structural image is further encoded into a structural latent:
\begin{equation}
\mathbf{z}_s = \mathcal{E}(I_s).
\end{equation}
During denoising, we concatenate the structural latent $\mathcal{E}(I_s)$ with the person latent $\mathcal{E}(I_p)$ along the channel dimension to obtain a structure-aware person latent $\tilde{\mathbf{z}}_p$.

A key advantage of SMPL-X lies in its disentangled representation of body shape and pose. By fixing the shape parameters while editing only the pose parameters, we can render structural conditions under arbitrary target poses while preserving the body shape of the source person. This enables pose-controllable virtual try-on without requiring additional template images.

\section{Experiments}

\begin{figure*}[t]
	\begin{center}
		\includegraphics[width=\linewidth]{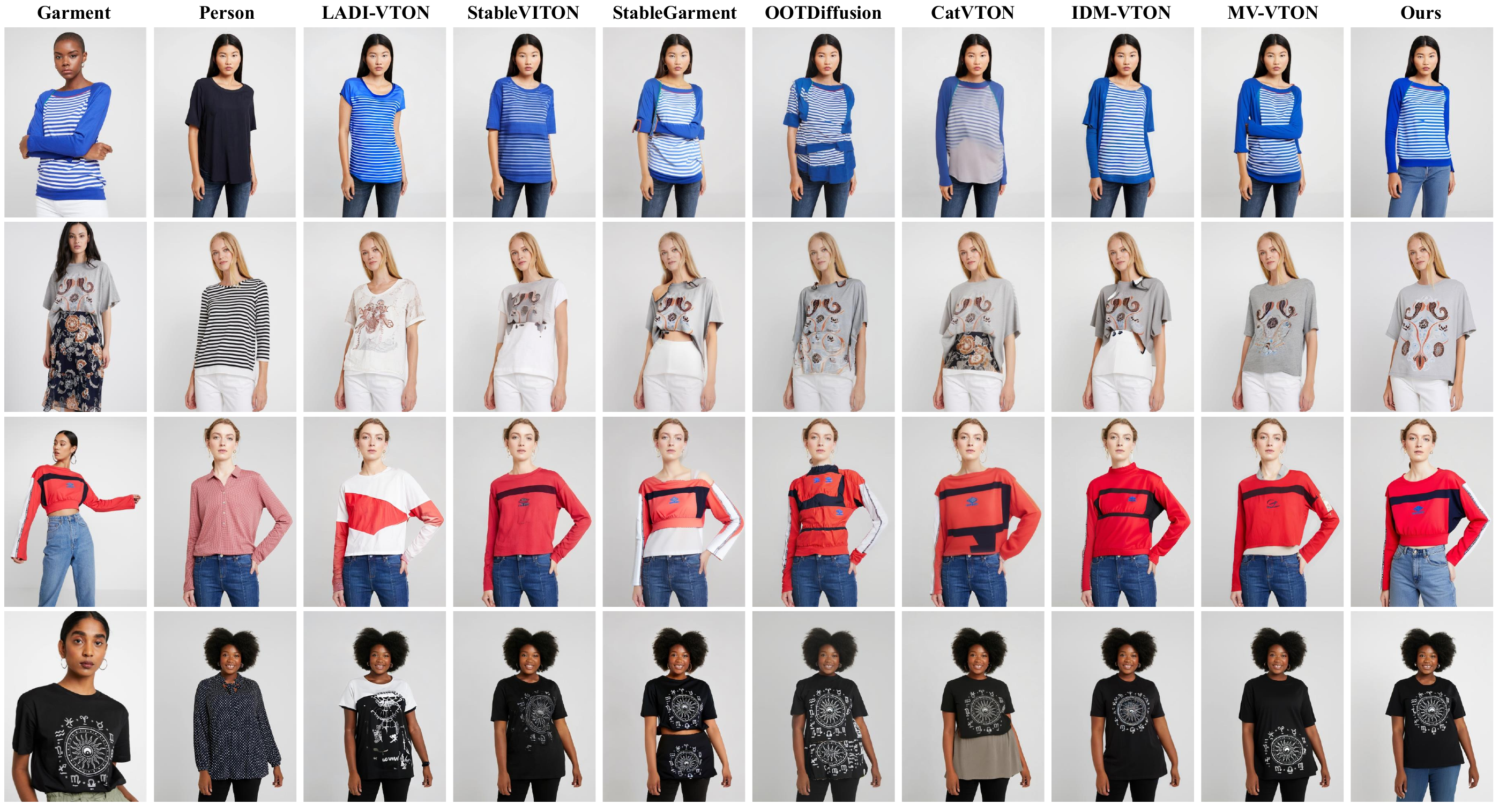} 
	\end{center}
    \vspace{-11pt}
    \caption{\textbf{Qualitative evaluation of virtual try-on on VITON-HD dataset.} OMFA shows a clear advantage in handling person-to-person virtual try-on.}
    \vspace{-5pt}
 \label{fig:vitonhd_tryon}
\end{figure*}

\begin{table*}[ht] 
\centering
\setlength{\tabcolsep}{5.0pt}
\renewcommand{\arraystretch}{1.0}%
    \caption{\textbf{Quantitative evaluation of virtual try-on on VITON-HD dataset.} The best and second-best results are demonstrated in \textbf{bold} and \underline{underlined}, respectively.} 
    \vspace{-5pt}
    \begin{tabular}{l|cccc|ccccc}
    \toprule 
    \multirow{2}{*}{\textbf{Methods}}  & \multicolumn{4}{c|}{\textbf{Paired}} & \multicolumn{5}{c}{\textbf{Unpaired}} \\ 
    \cline{2-10} 
                      & SSIM$\uparrow$ & LPIPS$\downarrow$ & FID$\downarrow$  & KID$\downarrow$  & CLIP-I$\uparrow$  & DINO$\uparrow$  & FID$\downarrow$ & KID$\downarrow$   & LLM$\uparrow$   \\ 
    \midrule 
    LADI-VTON~\cite{morelli2023ladi} & 0.856 & 0.137 & 12.072 & 5.763 & 0.821 & 0.810 & 15.925 &7.732 &7.52 \\ 
    StableGarment~\cite{wang2024stablegarment} & 0.827 & 0.115 & 12.463 & 6.311 & 0.805 & 0.799 & 18.539 & 10.508& 7.59 \\ 
    StableVITON~\cite{kim2024stableviton} & 0.849 & 0.123 & 9.566 & 3.062 & 0.819 & 0.800 & 12.489 & 4.124& 7.56 \\ 
    OOTDiffusion~\cite{xu2025ootdiffusion} & 0.827 & 0.107 & 8.242 & 1.895 & 0.831 & 0.809 & 12.817 & 3.909 & 7.80 \\ 
    IDM-VTON~\cite{choi2024improving} & 0.857 & \textbf{0.085} &  \underline{7.117} & 1.813 & 0.846 & 0.821 &  \underline{10.619} & \underline{2.703} & \underline{8.17} \\ 
    MV-VTON~\cite{wang2025mv}  & \textbf{0.879} & 0.114 & 7.980 & 2.760 & 0.841 &  0.827 &  11.002 &	2.906 & 8.07 \\ 
    CatVTON~\cite{chong2024catvton} & 0.843 & 0.099 & \textbf{7.074} & 1.890 & 0.833 & 0.817 & 12.178 & 3.987 & 7.98 \\ \midrule 
    CatVTON~\cite{chong2024catvton}+TryOffDiff~\cite{velioglu2024try} &  0.836 &  0.118 & 12.611 & 3.663 & 0.825 & 0.821 & 12.611 & 3.663 & - \\ 
    IDM-VTON~\cite{choi2024improving}+TryOffDiff~\cite{velioglu2024try} &  0.859 &  \underline{0.090} &  7.849 & \underline{1.312} & \underline{0.847} & \underline{0.830} & 11.299 & 2.670 & - \\ 
    MV-VTON~\cite{wang2025mv}+TryOffDiff~\cite{velioglu2024try} &  \textbf{0.879} &  0.114 & 8.082 & 2.857 & 0.840 & 0.827 & 11.174 & 3.288 & - \\
    Any2AnyTryon~\cite{Guo_2025_ICCV} & 0.861  &  0.151 & 7.168 & 1.732 & 0.829 & 0.818 & 13.398 & 5.401 & - \\
    \midrule 
    Ours &  \underline{0.862} &  0.098 & 7.170 & \textbf{1.160} & \textbf{0.876} & \textbf{0.850} & \textbf{10.527} & \textbf{1.923} & \textbf{8.32} \\ 
    \bottomrule 
    \end{tabular}
    \vspace{-10pt}
\label{tab:vitonhd_tryon} 
\end{table*}

\subsection{Experimental Setup}
\paragraph{Datasets.}
We train and evaluate our model on two publicly available fashion datasets: VITON-HD \cite{choi2021viton} and DeepFashion-MultiModal dataset \cite{liuLQWTcvpr16DeepFashion, jiang2022text2human, huang2024parts2whole}. VITON-HD contains 13,679 image pairs of frontal half-body models and corresponding upper-body garments, with 11,647 for training and 2,032 for testing. In the DeepFashion-MultiModal dataset, each sample includes not only images of person and garment but also a pair of target images in two poses. We select around 40,000 as training samples and 1,100 test samples. 
To prepare the inputs, we use a segmentation method \cite{li2020self} to obtain different image regions.
\vspace{-10pt}
\paragraph{Implementation Details.}
In our experiments, we initialize the models by inheriting the pretrained weights of Stable Diffusion XL, and finetune the parameters of the denoising UNet with the AdamW optimizer \cite{loshchilov2017decoupled}, using $\beta_1 = 0.9$ and $\beta_2 = 0.999$.
The model is trained at a high resolution of $768 \times 1024$ on 4 NVIDIA A800 GPUs for 65,000 steps, with a batch size of 8 and a learning rate of $1e^{-6}$.
To enable classifier-free guidance \cite{ho2022classifier} and maintain generation diversity, we randomly drop each conditional reference feature with a probability of 0.05. During inference, we adopt the DDIM sampler \cite{song2020denoising} with 50 diffusion steps and set the guidance scale to 2.0.

\vspace{-10pt}
\paragraph{Comparison Methods.}
For the try-on task, we compare our method with several state-of-the-art methods, including LADI-VTON \cite{morelli2023ladi}, StableGarment \cite{wang2024stablegarment}, StableVITON \cite{kim2024stableviton}, OOTDiffusion \cite{xu2025ootdiffusion}, IDM-VTON \cite{choi2024improving}, CatVTON \cite{chong2024catvton} and MV-VTON \cite{wang2025mv}. Under the more realistic setting where template garments are unavailable, we adapt the input pipelines of these methods to use segmented garment images.
For the multi-pose try-on task, we compare our method with the baseline IDM-VTON \cite{choi2024improving}. 
For the try-off task, we evaluate our method against two recent approaches: TryOffDiff \cite{velioglu2024try} and TryoffAnyone \cite{xarchakos2024tryoffanyone}.  
Additionally, to evaluate cross-task compatibility, we consider five two-stage pipelines that perform garment synthesis followed by try-on.
Our method and Any2AnyTryon \cite{Guo_2025_ICCV} both follow this two-stage design. We also include three additional combinations that feed TryOnDiff-generated garments into CatVTON \cite{chong2024catvton}, IDM-VTON \cite{choi2024improving}, and MV-VTON \cite{wang2025mv}.
We use the pre-trained checkpoints provided in the official repositories of all compared methods.

\vspace{-10pt}
\paragraph{Evaluation Metrics.}
For paired settings, we evaluate similarity between synthesized and ground-truth images using SSIM \cite{wang2004image}, LPIPS \cite{zhang2018unreasonable}, FID \cite{heusel2017gans}, and KID \cite{binkowski2018demystifying}. For unpaired settings, in addition to computing FID and KID, we further calculate CLIP-I \cite{radford2021learning} and DINO \cite{oquab2023dinov2} similarity between the segmented garment region and the corresponding reference garment to assess garment-level semantic consistency. To ensure a fair comparison with mask-based methods, we use the agnostic map to preserve the unedited regions, following CatVTON \cite{chong2024catvton}.
Moreover, given the person and garment images, we use GPT-4o-mini to provide a comprehensive score for the try-on result. The score ranges from 0 to 10. For the garment generation task, we additionally report DISTS \cite{ding2020image}, a perceptual similarity metric designed to capture both structural and textural fidelity between the generated garment image and the ground truth.

\begin{figure}[t]
\centering
\includegraphics[width=\linewidth]{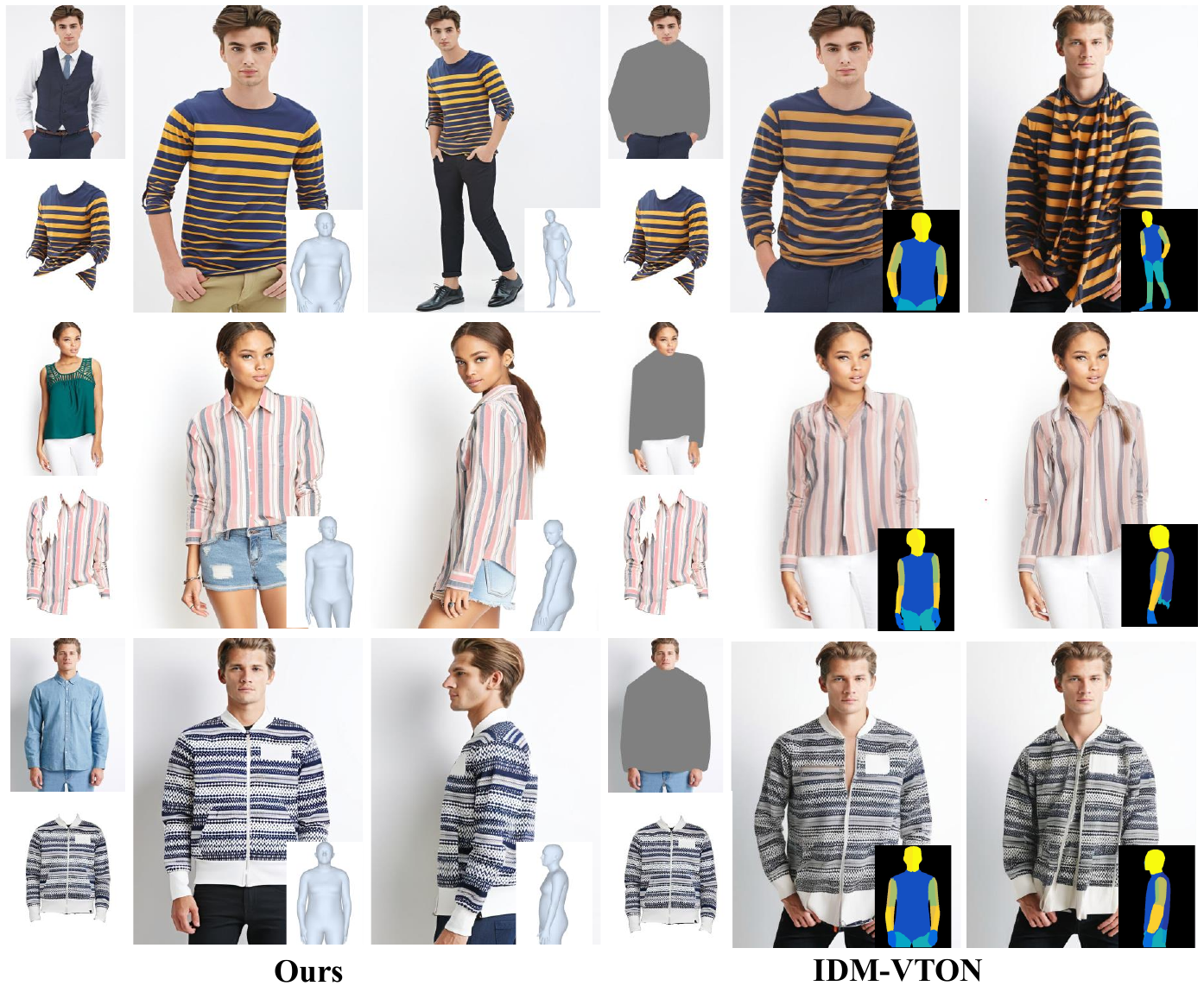} 
    \vspace{-15pt}
    \caption{\textbf{Qualitative comparison of multi-pose try-on results with IDM-VTON on DeepFashion-MultiModal.} To adapt the input of IDM-VTON, we keep the agnostic mask unchanged and replace the input DensePose representation with the target pose to investigate its capability for pose transfer.}
    \vspace{-5pt}
    \label{fig:deepfashion_tryon}
\end{figure}

\begin{figure}[ht]  
	\centering
    \includegraphics[width=\linewidth]{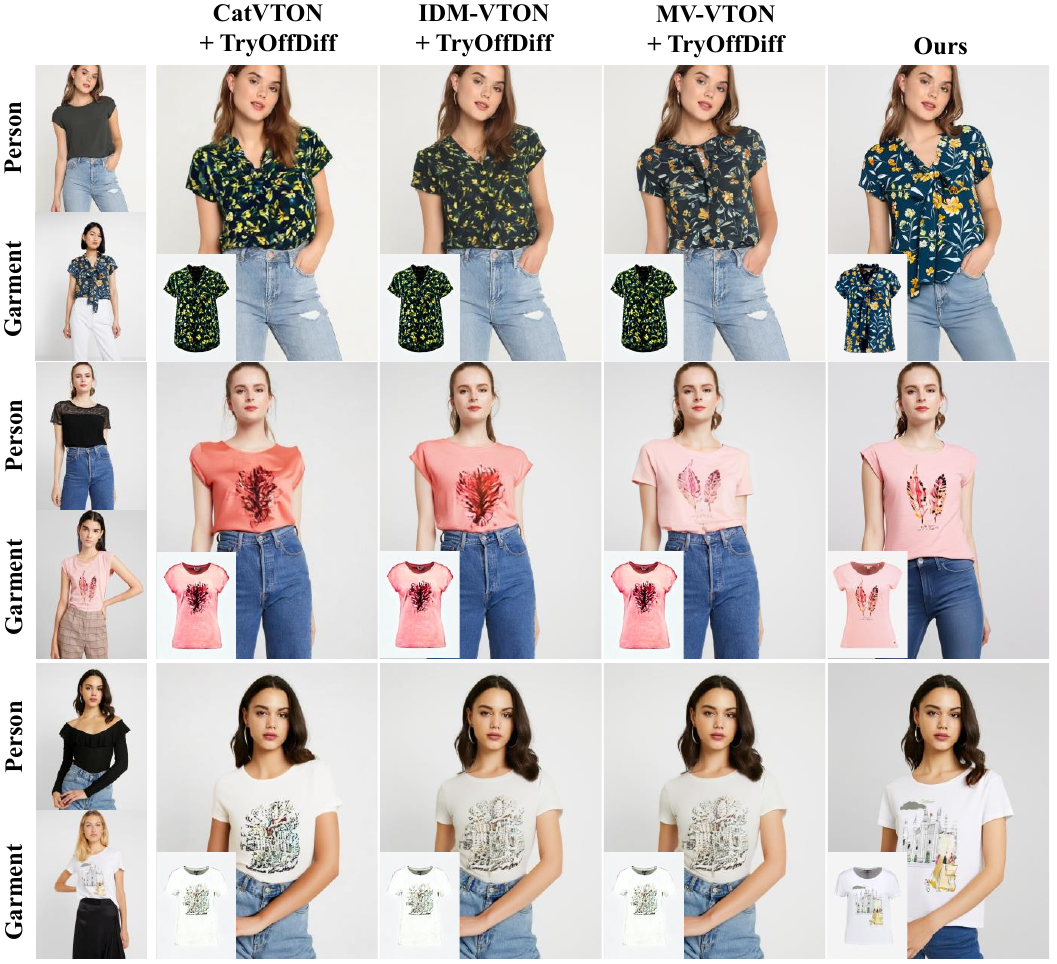}
    \vspace{-15pt}
    \caption{\textbf{Qualitative comparison of TryOffDiff-combined try-on pipelines and our unified framework.} Methods combined with TryOffDiff tend to blur patterns, whereas our method better preserves garment details.}
    \vspace{-15pt}
    \label{fig:tryoff+tryon}
\end{figure}

\subsection{Virtual Try-on}
\begin{figure}[ht]  
	\centering
    \vspace{-10pt}
    \includegraphics[width=0.97\linewidth]{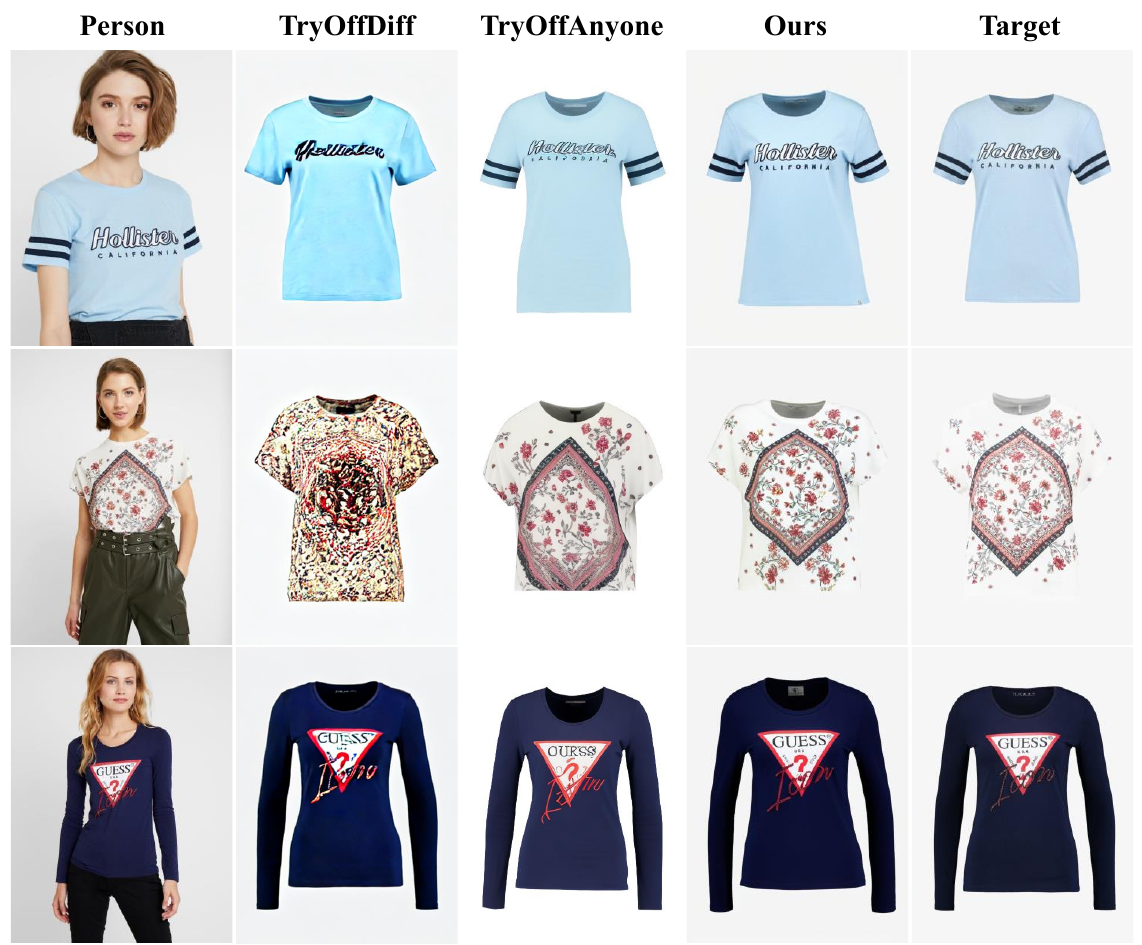}
    \vspace{-5pt}
    \caption{\textbf{Qualitative evaluation of virtual try-off on VITON-HD dataset.} OMFA successfully reconstructs clear patterns and text of the garment.}
    \vspace{-7pt}
    \label{fig:garment}
\end{figure}

\vspace{-5pt}
\paragraph{Person-to-person Virtual Try-on.}
Tab.~\ref{tab:vitonhd_tryon} reports the virtual try-on results on the VITON-HD dataset. 
In the paired setting, our method achieves comparable overall metrics. While some baselines report slightly higher SSIM scores, this is most likely due to the input warped cloth being well aligned with the target, making it easier to maintain garment appearance. 
When garments recovered by existing try-off methods are used as inputs to the try-on framework (see Fig.~\ref{fig:tryoff+tryon} and Fig.~\ref{fig:garment}), fine-grained texture details remain inadequately recovered.
Although such inputs usually provide a complete garment silhouette, the subsequent try-on stage still exhibits degraded fabric realism or blurred patterns/LOGOs. Benefiting from the ability to reconstruct garments, our method performs notably better in the more challenging unpaired try-on setting, particularly on CLIP-I and DINO similarity. In Fig.~\ref{fig:vitonhd_tryon}, we present a qualitative comparison between our method and multiple advanced approaches on the VITON-HD dataset, highlighting its distinct advantage in the person-to-person try-on scenario.

\vspace{-10pt}
\paragraph{Multi-pose Virtual Try-on.}
We further explore the multi-pose try-on task. Tab.~\ref{tab:deepfashion_tryon} shows that OMFA outperforms the baseline across all metrics, confirming its flexibility with respect to pose and view variations. As shown in Fig.~\ref{fig:deepfashion_tryon}, the pose of the image generated by IDM-VTON is primarily determined by the unmasked regions, and inconsistent pose inputs lead to incorrect garment deformations. In contrast, our mask-free method leverages 3D human representations during generation, enabling more flexible pose transfer and size-aware garment fitting.
\setlength{\tabcolsep}{3.5pt}
\begin{table}[htbp]
    \small
    \centering
    \renewcommand{\arraystretch}{1.0}
    \caption{\textbf{Quantitative evaluation} of multi-pose try-on results on the DeepFashion-MultiModal dataset.}
    \vspace{-7pt}
    \begin{tabular*}{0.9\columnwidth}{@{\extracolsep{\fill}}c|ccccc@{}}
    \toprule 
    \textbf{Methods} & LPIPS$\downarrow$ & FID$\downarrow$ & KID$\downarrow$ & CLIP-I$\uparrow$ & DINO$\uparrow$ \\ \midrule 
    IDM-VTON    &   0.163    &  18.134  &  4.326   &  0.857  &   0.833   \\ 
    Ours        &   \textbf{0.124}    &  \textbf{14.548}  &  \textbf{3.121}   &  \textbf{0.879}  &   \textbf{0.857}
   \\ \bottomrule
    \end{tabular*}
    \vspace{-7pt}
    \label{tab:deepfashion_tryon}
\end{table}

\vspace{-4pt}
\subsection{Virtual Try-off}
Tab.~\ref{tab:tryoff-Quantitative} shows quantitative results for virtual try-off, where our method outperforms advanced approaches across all five metrics, showing significant advantages in detail preservation, structural and textural consistency, and semantic alignment.
Fig.~\ref{fig:garment} presents a comparison of garment reconstruction results on the VITON-HD dataset between our method and other try-off approaches. Specifically, TryOffDiff mainly recovers coarse shape and color and often misses fine patterns, while TryOffAnyone better handles complex patterns but still blurs or omits text. By comparison, our method shows clear and consistent advantages in detail preservation, particularly in the clarity of textual contours and pattern boundaries.

\vspace{-5pt}
\setlength{\tabcolsep}{2pt}
\begin{table}[ht]
    \small
    \centering
    \renewcommand{\arraystretch}{1.2}
    \caption{\textbf{Quantitative evaluation} of virtual try-off on the VITON-HD dataset.}
    \vspace{-8pt}
    \begin{tabular*}{0.9\columnwidth}{@{\extracolsep{\fill}}c|ccccc@{}}
    \hline
        \textbf{Methods} & LPIPS$\downarrow$ & DISTS$\downarrow$ & CLIP-I$\uparrow$ & KID$\downarrow$ & FID$\downarrow$ \\ \hline
        TryOffDiff & 0.323 & 0.237 & 0.902 & 7.912 & 21.919 \\ 
        TryOffAnyone & 0.269 & 0.217 & 0.923 & 2.533 & 12.453 \\ 
        Ours & \textbf{0.225} & \textbf{0.192} & \textbf{0.944} & \textbf{1.495} & \textbf{9.121} \\ \hline
    \end{tabular*}
    \vspace{-10pt}
    \label{tab:tryoff-Quantitative}
\end{table}

\subsection{Ablation Studies}

\begin{figure} [ht]
	\centering
    \includegraphics[width=\linewidth]{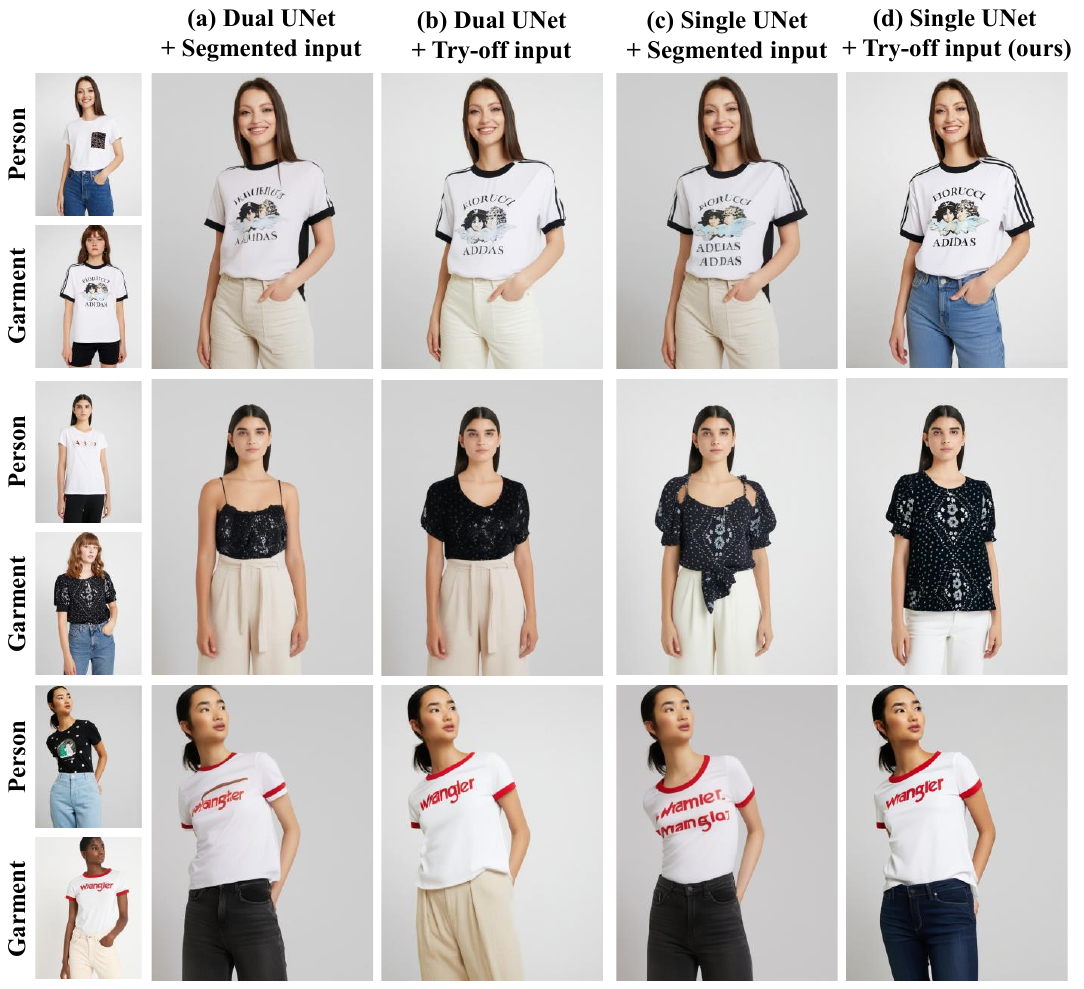}  
    \vspace{-15pt}
    \caption{\textbf{Visual comparisons of our proposed modules.} We also compare variants using a segmented input versus a try-off input. }
    \vspace{-7pt}
    \label{fig:ablation}
\end{figure}

\paragraph{Effectiveness of the Bidirectional Tweedie Diffusion.}
In the baseline setting, we follow IDM-VTON \cite{choi2024improving} and train parallel UNets with a ReferenceNet that encodes the garment images and injects their features into the denoising UNet. We then replace ReferenceNet with a single UNet and adopt the Bidirectional Tweedie Diffusion to process the spatially joint input. As shown in Fig.~\ref{fig:ablation}(c), even with a warped garment input, this mechanism delivers more faithful details compared to (a). The quantitative results in the first group (lines 1 and 3) and the second group (lines 2 and 4) of Tab.~\ref{tab:ablation} further demonstrate improved visual fidelity. 

\setlength{\tabcolsep}{1.2pt}
\begin{table}[t]
    \small
    \centering
    \caption{\textbf{Ablation study of the proposed modules on the VITON-HD dataset.} We use the Dual UNet architecture as the baseline, where “Single UNet” denotes our LLM-inspired bidirectional diffusion design. Our OMFA achieves consistently superior scores on all evaluation metrics.}
    \vspace{-5pt}
    \begin{tabular*}{0.92\columnwidth}
      {@{\extracolsep{\fill}} cc|cc|ccc@{}}
      \toprule
      \multirow{2}{*}{\textbf{Arch.}} &
      \multirow{2}{*}{\textbf{\shortstack{Garment\\Input}}} &
      \multicolumn{2}{c}{\textbf{Paired}} &
      \multicolumn{3}{c}{\textbf{Unpaired}} \\
      \cmidrule(lr){3-4}\cmidrule(lr){5-7}
      & & SSIM$\uparrow$ & LPIPS$\downarrow$
        & FID$\downarrow$ & KID$\downarrow$ & CLIP-I$\uparrow$ \\
      \midrule
      Dual UNet & Segmented & 0.839 & 0.160 & 19.139 & 9.841 & 0.805 \\
      Dual UNet & Try-off & 0.843 & 0.148 & 13.587 & 4.707 & 0.856 \\
      Single UNet & Segmented & 0.853 & 0.098 & 12.663 & 3.875 & 0.860 \\
      Single UNet & Try-off & \textbf{0.862} & \textbf{0.098} & \textbf{10.527} & \textbf{1.920} & \textbf{0.876} \\
      \bottomrule
    \end{tabular*}
    \vspace{-7pt}
    \label{tab:ablation}
\end{table}

\vspace{-10pt}
\paragraph{Effectiveness of the unified generation strategy.}
We further validate the unified generation strategy for try-on and try-off in person-to-person scenarios. When the exemplar garment is unavailable, try-on results with segmented-garment input may exhibit incomplete contours and occluded details (see Fig.~\ref{fig:ablation}(a) and (c)). In contrast, our unified pipeline first performs try-off and then conducts garment transfer within a single model, producing the most complete and visually consistent results . The quantitative results in the two ablation groups (lines 1 and 2, and lines 3 and 4 in Tab.~\ref{tab:ablation}) further support this conclusion.

\vspace{-10pt}

\begin{figure}[ht]
	\centering
    \includegraphics[width=\linewidth]{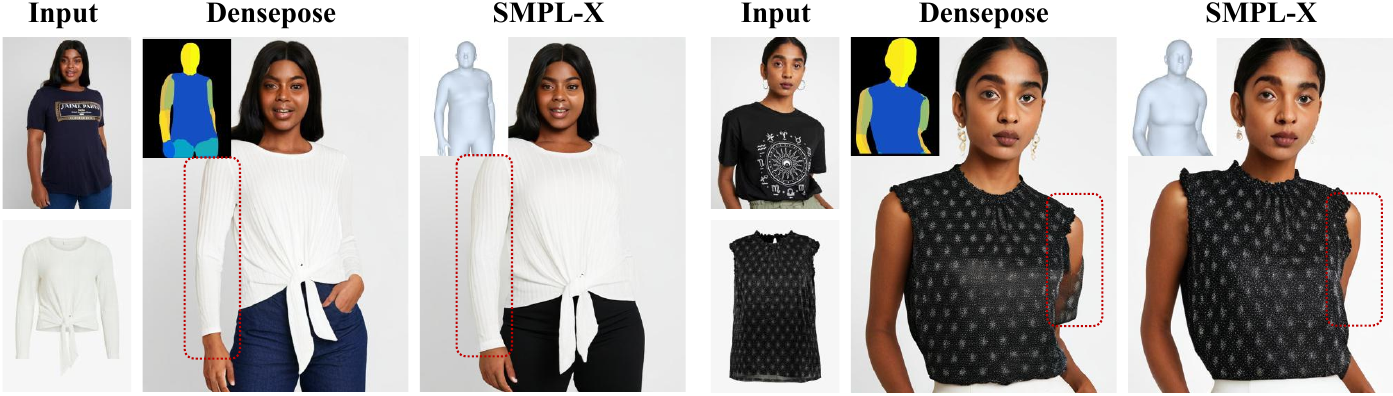}  
    \caption{\textbf{Qualitative ablation of structural conditioning} on VITON-HD, illustrating the benefit of SMPL-X over DensePose.}
    \vspace{-8pt}
    \label{fig:ablation-smplx}
\end{figure}

\paragraph{Impact of SMPL-X-based conditioning.}
We additonally train two small-resolution variants on VITON-HD, respectively using DensePose and SMPL-X as the body-structure inputs. As shown in Tab.~\ref{tab:ablation-smplx}, SMPL-X performs similarly to DensePose. However, as illustrated in Fig.~\ref{fig:ablation-smplx}, DensePose sometimes causes shape distortion (e.g., arms looking thinner; left) and boundary artifacts (right). By contrast, SMPL-X offers explicit geometric modeling, allowing direct control of body pose and shape through its parameters.

\setlength{\tabcolsep}{2pt}
\begin{table}[ht]
    \small
    \centering
    \caption{\textbf{Quantitative ablation of structural conditioning} on VITON-HD: DensePose vs. SMPL-X.}
    \vspace{-5pt}
    \begin{tabular*}{0.9\columnwidth}{@{\extracolsep{\fill}}c|ccccc@{}}
    \hline
    \renewcommand{\arraystretch}{3.0}
    \textbf{\shortstack{Structural\\ Condition}} & SSIM$\uparrow$ & LPIPS$\downarrow$ & CLIP-I$\uparrow$ & KID$\downarrow$ & FID$\downarrow$ \\ \hline
    \renewcommand{\arraystretch}{1.8}
    \textbf{Densepose} & 0.839 & \textbf{0.132} & 0.863 & 11.645 & 3.610 \\ 
    \textbf{SMPL-X} & \textbf{0.839} & 0.152 & \textbf{0.872} & \textbf{11.781} & \textbf{3.602} \\ \hline
    \end{tabular*}
    \vspace{-15pt}
    \label{tab:ablation-smplx}
\end{table}

\section{Conclusion}
We presented \textbf{OMFA}, a unified diffusion-based framework for virtual try-on and try-off that overcomes key limitations of prior methods, including dependence on garment templates, segmentation masks, and fixed poses. OMFA introduces a novel \emph{bidirectional Tweedie diffusion} mechanism for efficient, interactive garment-person transformation with fine-grained subtask control. It operates in a fully mask-free manner and requires only a single portrait and a target pose, making it practical for real-world use. With SMPL-X–based pose conditioning, OMFA enables flexible, multi-view try-on from a single image. Extensive experiments confirm its effectiveness and generalizability across both tasks.

\section{Acknowledgments}

This work was supported in part by National High-Level Young Talent Program (Grant 2025HY00260104), in part by the Fundamental Research Funds for Higher Education Institutions allocated to Sun Yat-sen University (Grants 25hytd007 and 2025RGZN009), in part by the Guangdong Provincial High-Level Young Talent Program (Grant 2025HYSPT0707), in part by the Tuoyuna Grant (HT-99982025-0564), in part by the Faculty Start-up Research Fund (Grant 67000-12255002), and in part by the Huawei Strategic Research Institute Talent Fund.
{
    \small
    \bibliographystyle{ieeenat_fullname}
    \bibliography{main}
}

\clearpage
\setcounter{page}{1}
\maketitlesupplementary

\setcounter{secnumdepth}{2}
\section{Implementation Details}
\subsection{Training and Inference Details}
We first train the model on the VITON-HD dataset with a resolution of $384 \times 512$ for 20K iterations. Then, keeping the same learning rate and batch size, we finetune the model on both the VITON-HD and DeepFashion-MultiModal datasets using a higher resolution of $768 \times 1024$. For data augmentation, we enhance the background color of the generated garment. Specifically, we use a tensor of the same size as the input garment, with all values set to 255, and concatenate it with the garment latent $\mathcal{E}(I_g)$ along the channel dimension. To align the latent with the UNet input along the channel dimension, we apply separate convolutional layers to each component of the joint input, projecting their channels to 320. Each convolutional layer is initialized with the first several channels of the corresponding layer in the pretrained UNet. 
During inference, if the try-on task is required, the inputs needed are the person image $I_p$, the garment $I_g$, and the person's portrait $I_h$, whereas for the try-off task, only $I_p$ is needed, with the other inputs set to 0.
Our implementation is based on the PyTorch deep learning framework (version 2.1.2), with the diffusion model adapted from HuggingFace's Diffusers library.

\subsection{Evaluation Metrics}
\begin{figure*}[t]
	\begin{center}
		\includegraphics[width=\linewidth]{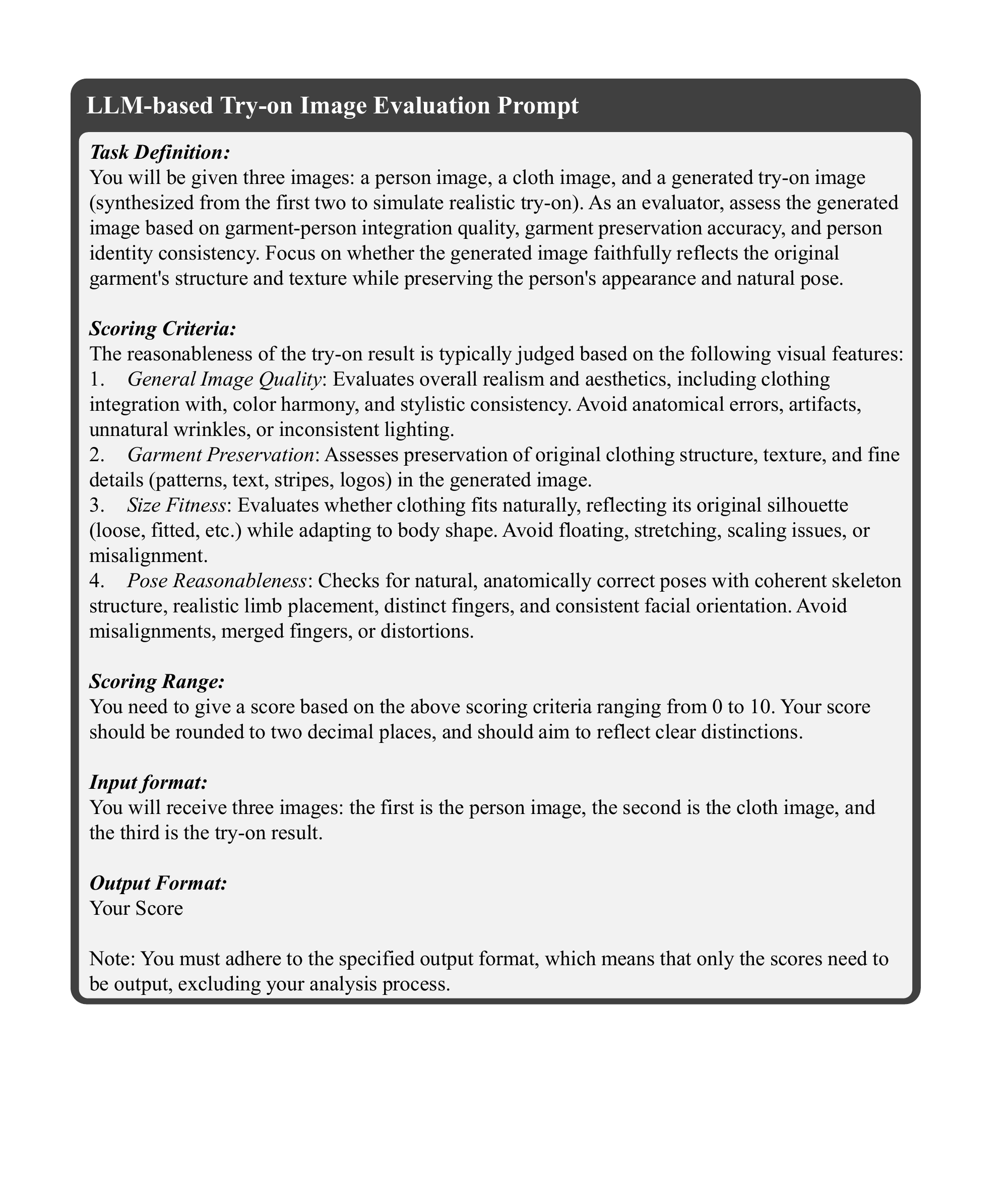} 
	\end{center}
    \caption{\textbf{The prompt for GPT-4o-mini to evaluate try-on results quality.}}
 \label{fig:llm_prompt}
\end{figure*}

In our experiments, we adopt a variety of evaluation metrics commonly used in generative tasks. Among them, SSIM \cite{wang2004image}, LPIPS \cite{zhang2018unreasonable}, KID \cite{binkowski2018demystifying}, and FID \cite{heusel2017gans} are widely used general metrics in related work. This section focuses on the detailed computation of several additional quantitative metrics used in our method, including DINO similarity \cite{oquab2023dinov2}, CLIP-I \cite{radford2021learning}, and LLM-based Image Scoring.
\paragraph{CLIP-I.}
CLIP focuses on semantic alignment similarity between images. Specifically, we utilize the CLIP-ViT-B/32 model as the feature extractor. Given a pair of images, the model encodes them into two 512-dimensional feature vectors. We then compute the cosine similarity between these vectors to measure their semantic similarity—a higher similarity indicates more semantically consistent content.
\paragraph{DINO Similarity.}
DINO similarity focuses on structural and fine-grained detail similarity between images. We utilize the DINOv2-Base model to extract features for a pair of images. For each image, we apply mean pooling across the patch embeddings from the final layer of the model's output, resulting in a 768-dimensional feature vector. We then calculate the cosine similarity to measure. 
\paragraph{LLM-based Image Scoring.}
We provide the model image, garment image, and the try-on result as input to GPT-4o-mini. The prompt for the multimodal large language model is presented in Fig.~\ref{fig:llm_prompt}.

\section{Additional Ablation Studies}
\paragraph{Experimental Setup for SMPL-X Ablation.}
Specifically, we trained two low-resolution models on VITON-HD, using DensePose (provided by the VITON-HD preprocessing) and SMPL-X respectively as structural conditioning, which were concatenated along the input channels. The configuration uses a resolution of 384×512, a per-GPU batch size of 8, and two 80-GB A100 GPUs for 45,000 training steps. At inference, the guidance scale is set to 2.0.

\paragraph{Ablation Studies for Joint Training Strategy.}
To evaluate the impact of unified training on performance, we conduct an ablation using the same low-resolution configuration as described above. In the joint-training setting, each batch is duplicated and divided into two halves, one for try-off training and the other for try-on training. Tab. ~\ref{tab:train-strategy} compares two training setups with no performance degradation. We attribute this to our bidirectional Tweedie diffusion mechanism, which explicitly defines the context and generation task by adjusting the combination of noising targets and conditions, thereby avoiding cross-task context confusion. Moreover, unified training encourages the model to learn bidirectional, shared feature correlations between garments and the human body, yielding more essential representations and improving the performance and robustness of both tasks. 

\begin{table}[!h]
    \centering
    \small
    \caption{\textbf{Quantity ablation of unified training strategy.} where  “(u)” indicates that the metric is computed in the unpaired setting. Training the network to learn both try-on and try-off tasks does not degrade performance. }
    \setlength{\tabcolsep}{2pt}
    \begin{tabular}{l|ccccc}
    \toprule
    \textbf{Setting} & SSIM$\uparrow$ & LPIPS$\downarrow$ & CLIP-I (u)$\uparrow$ & FID (u)$\downarrow$ & KID (u)$\downarrow$ \\
    \midrule
    Try-on training & \textbf{0.840} & \textbf{0.150} & 0.866 & 11.919 & 3.603 \\
    Joint training  & 0.839 & 0.152 & \textbf{0.872} & \textbf{11.781} & \textbf{3.602} \\
    \bottomrule
    \end{tabular}
    \label{tab:train-strategy}
\end{table}
\section{More Qualitative Results}
\subsection{Virtual Try-on}
\begin{figure*}[t]
	\begin{center}
		\includegraphics[width=\linewidth]{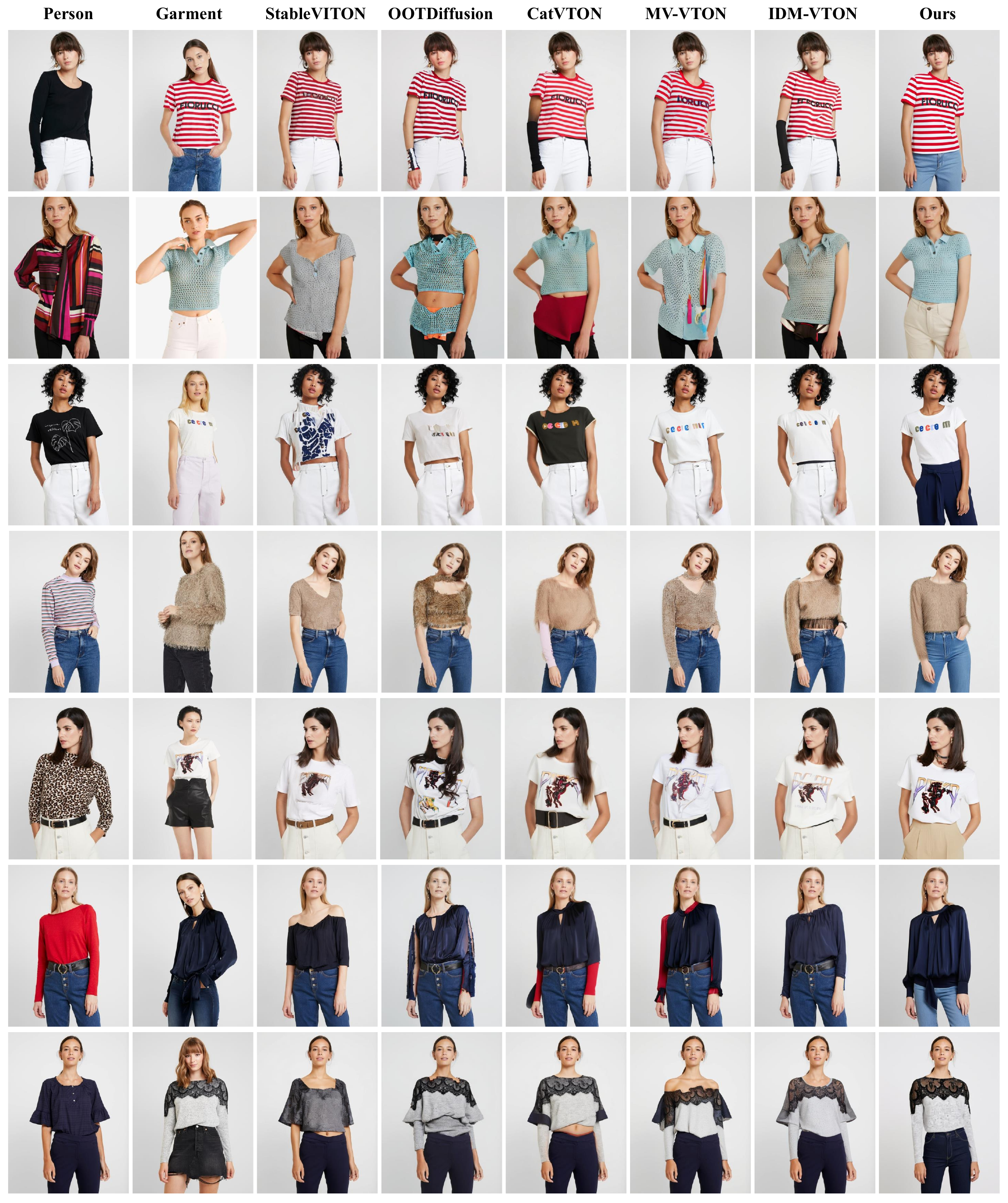} 
	\end{center}
    \caption{\textbf{More qualitative comparison of try-on results on VITON-HD.} \emph{Please zoom in to better observe the details.}}
 \label{fig:vitonhd_tryon2}
\end{figure*}

\begin{figure*}[t]
	\begin{center}
		\includegraphics[width=0.9\linewidth]{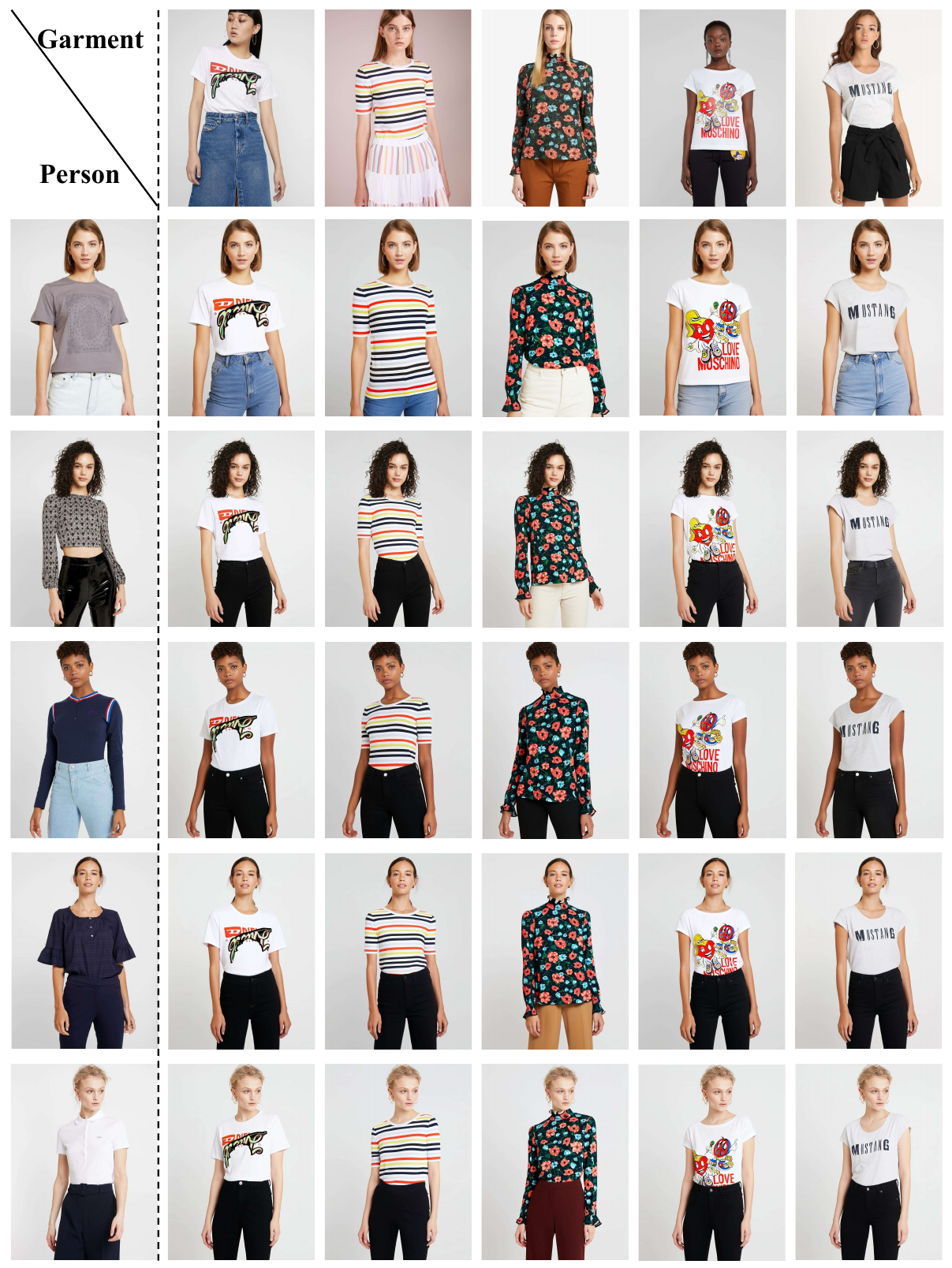} 
	\end{center}
    \caption{\textbf{More qualitative results on VITON-HD.} \emph{Please zoom in to better observe the details.}}
 \label{fig:vitonhd_ours1}
\end{figure*}

\begin{figure*}[t]
	\begin{center}
		\includegraphics[width=0.9\linewidth]{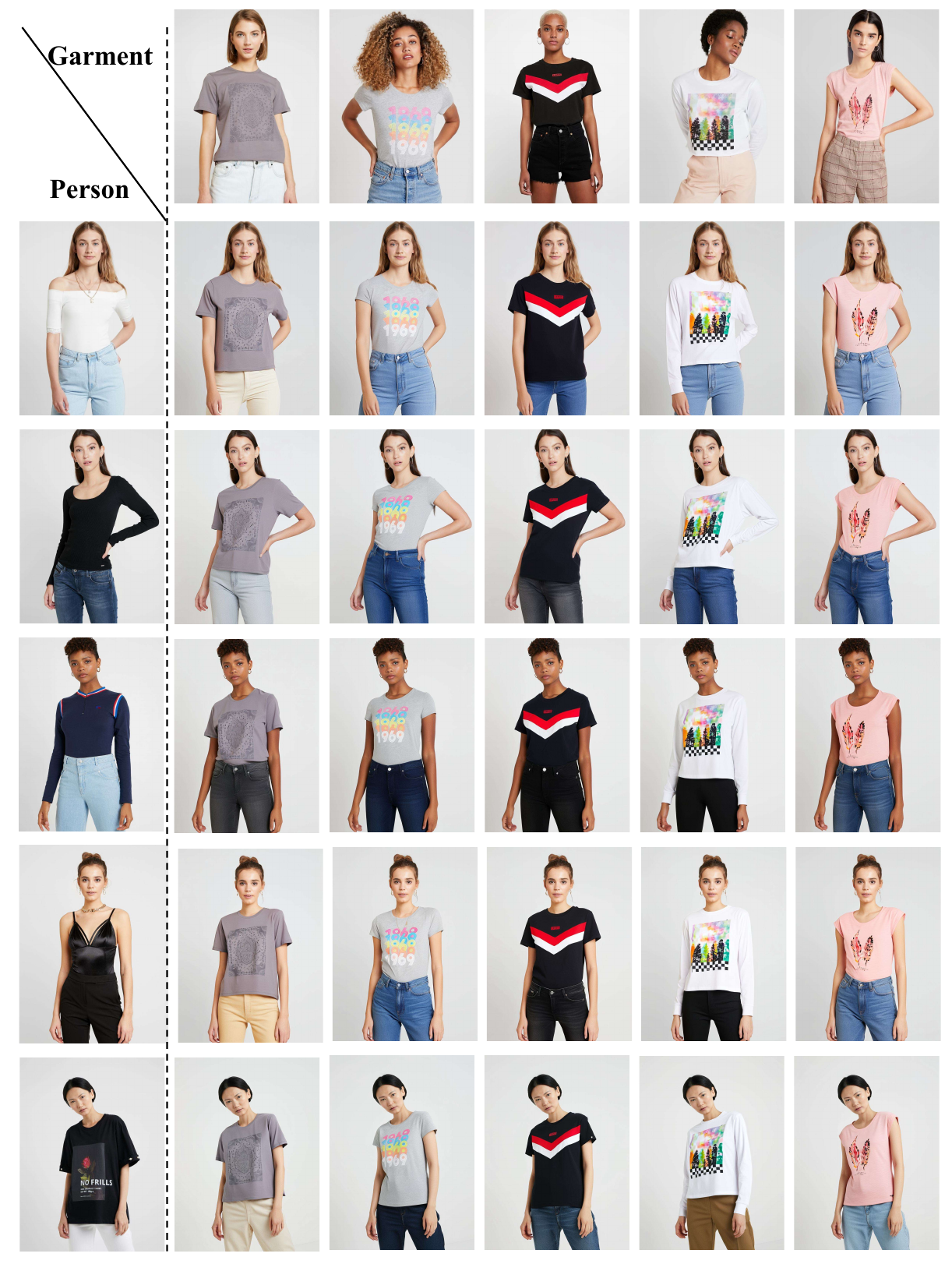} 
	\end{center}
    \caption{\textbf{More qualitative results on VITON-HD.} \emph{Please zoom in to better observe the details.}}
 \label{fig:vitonhd_ours2}
\end{figure*}

\paragraph{Person-to-person Virtual Try-on.}
Fig.~\ref{fig:vitonhd_tryon2} presents additional try-on comparative results in the person-to-person scenario on the VITON-HD dataset. Specifically, when the input clothing is not an exhibition garment, the input warp cloth often exhibits incomplete contours and distorted textures, which further exacerbates the artifacts in the try-on results, ultimately leading to suboptimal performance. 
Fig.~\ref{fig:vitonhd_ours1} and Fig.~\ref{fig:vitonhd_ours2}  present more generated results of our proposed OMFA model in this task, further demonstrating that our model maintains excellent detail preservation in both the try-off and try-on steps, leading to robust and high-fidelity results.

\begin{figure*}[t]
	\begin{center}
		\includegraphics[width=0.9\linewidth]{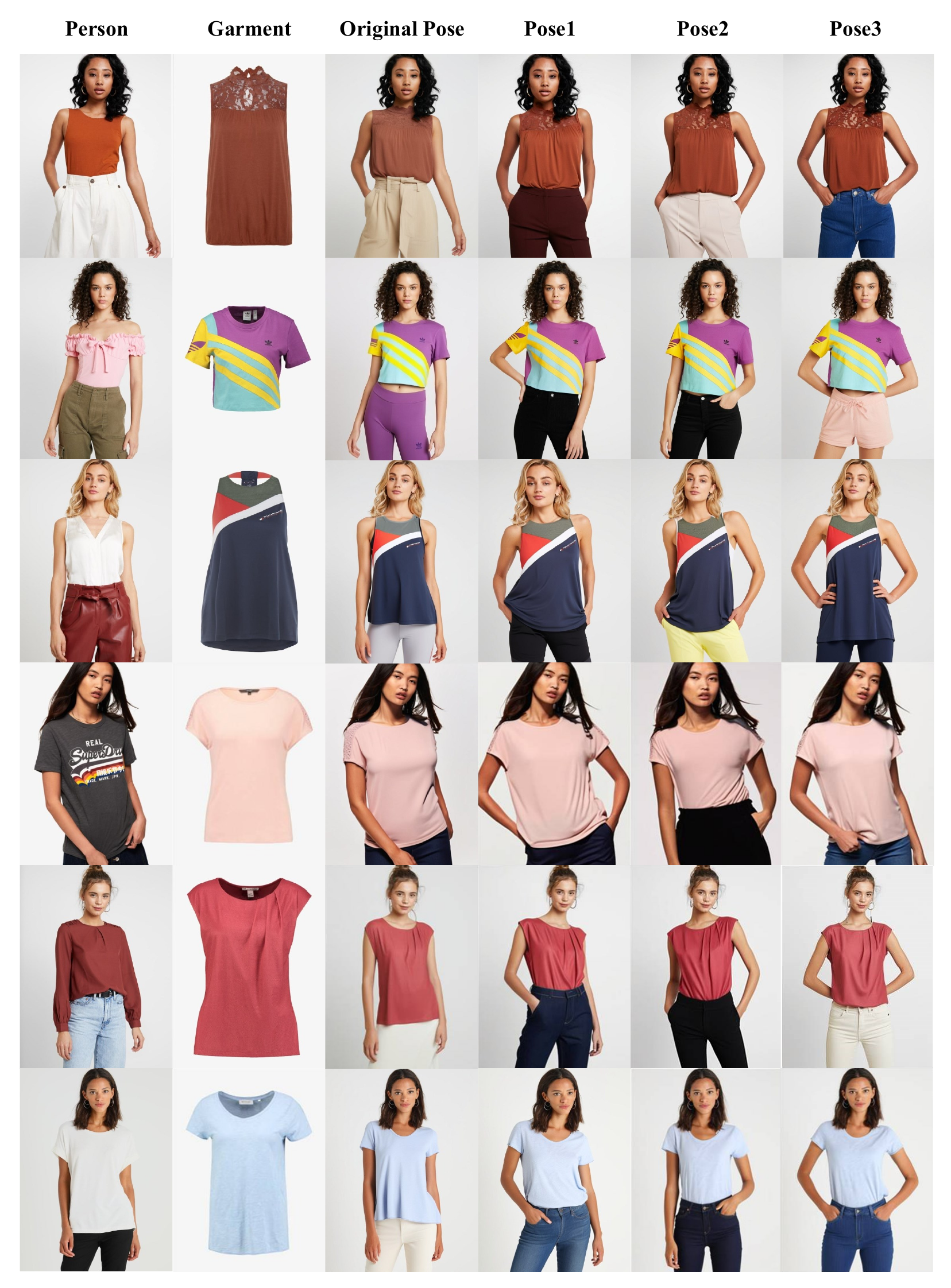} 
	\end{center}
    \caption{\textbf{Qualitative result of multi-pose try-on on VITON-HD.} \emph{Please zoom in to better observe the details.}}
 \label{fig:vitonhd_manypose}
\end{figure*}

\paragraph{Multi-pose Virtual Try-on.}
As shown in Fig.~\ref{fig:vitonhd_manypose}, we present pose transfer try-on results on the VITON-HD dataset. Specifically, we select three different target poses and replace the original pose parameters with the corresponding SMPL-X parameters to enable pose variation in try-on. We also provide try-on results under the original pose as a reference. Since we only replaced the pose parameters of SMPL-X while keeping the shape parameters unchanged, the generated body meshes exhibit different poses but consistent body shape, which helps achieve natural and identity-consistent try-on results. 

\subsection{Virtual Try-off}
\begin{figure*}[t]
	\begin{center}
		\includegraphics[width=\linewidth]{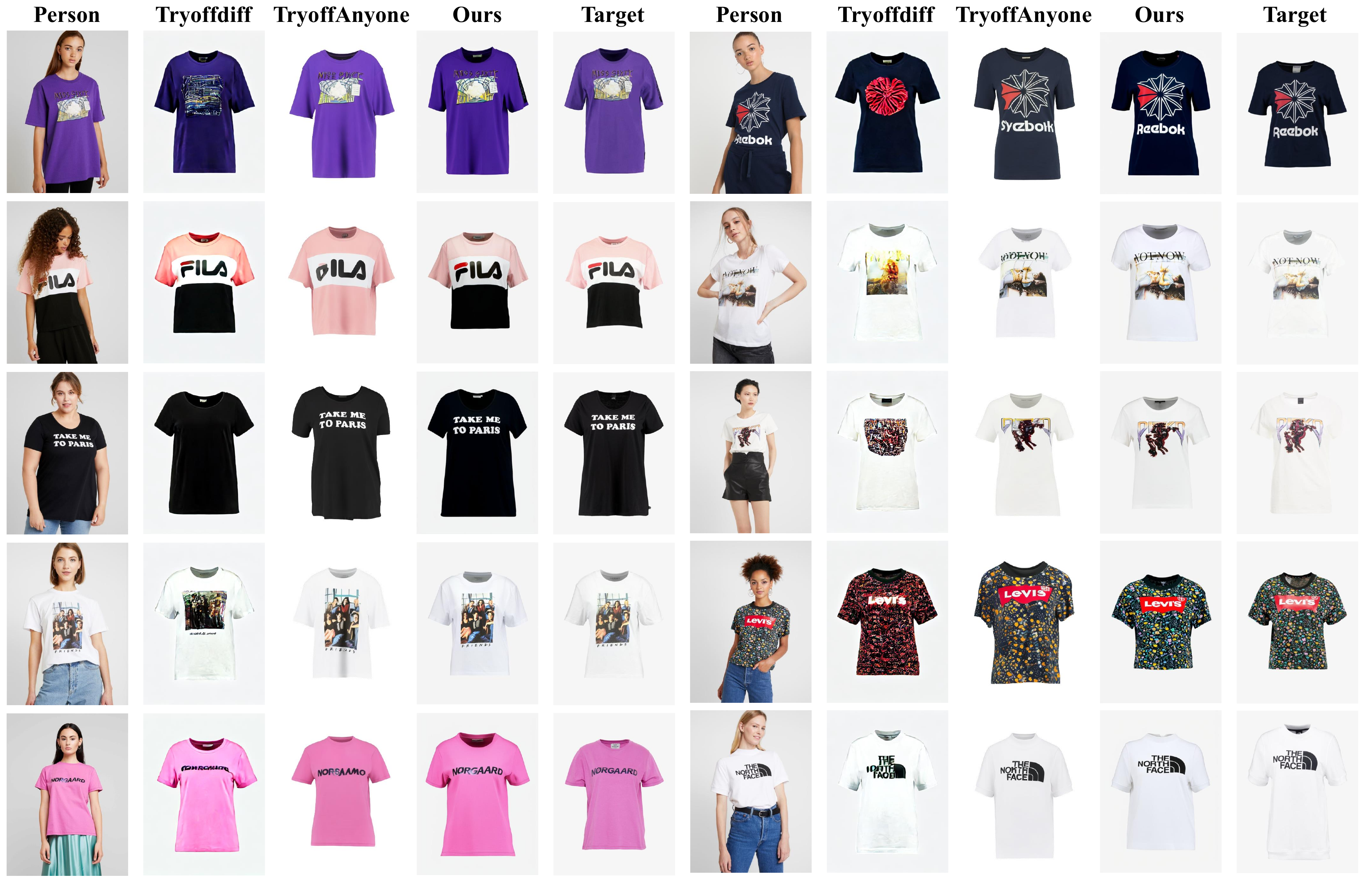} 
	\end{center}
    \caption{\textbf{More qualitative comparison of try-off results on VITON-HD.} \emph{Please zoom in to better observe the details.}}
 \label{fig:garment2}
\end{figure*}

\begin{figure*}[t]
	\begin{center}
		\includegraphics[width=\linewidth]{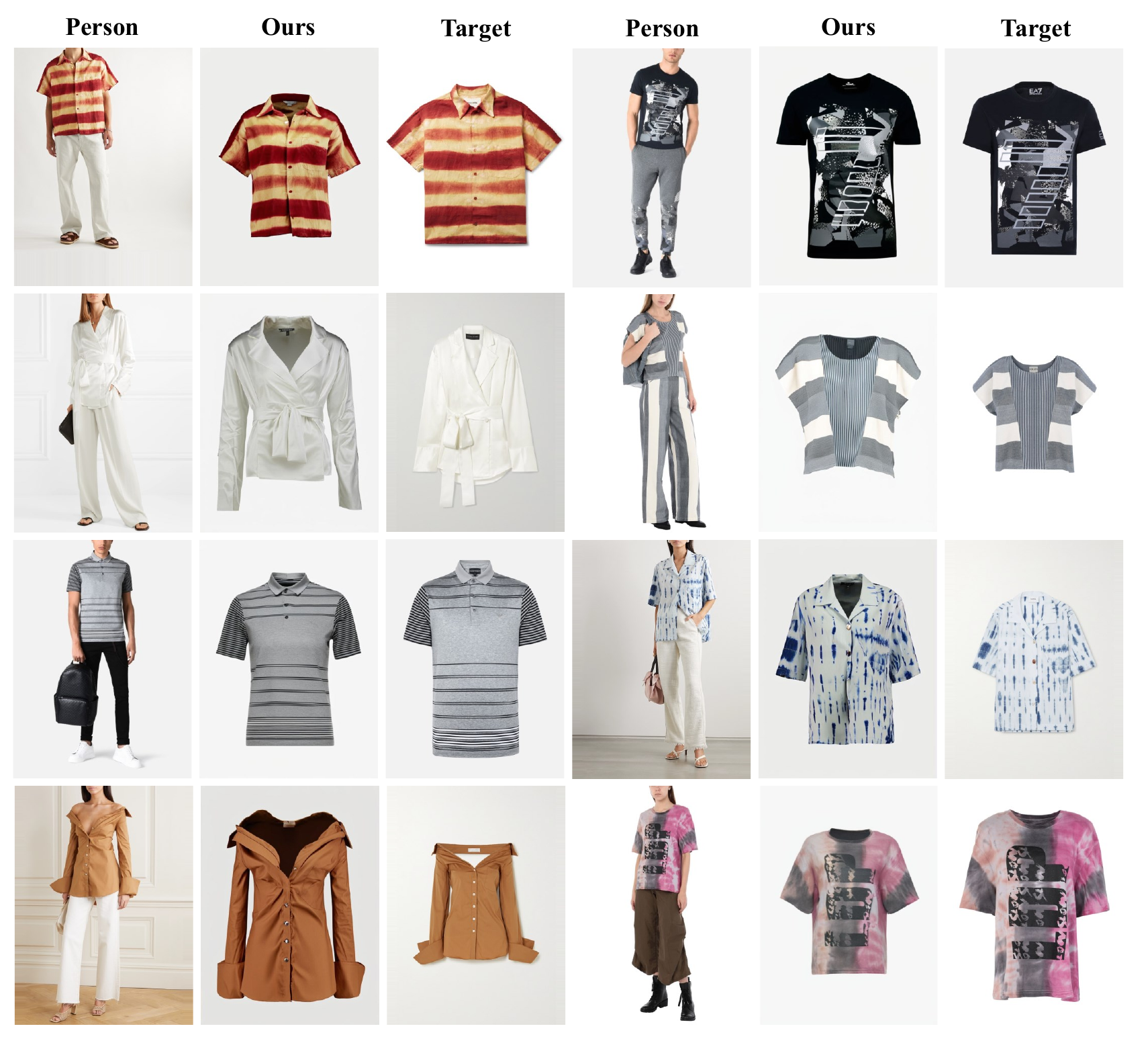} 
	\end{center}
    \caption{\textbf{Qualitative try-off results on DressCode upperbody.} \emph{Please zoom in to better observe the details.}}
 \label{fig:garment_dresscode}
\end{figure*}

\begin{figure*}[t]
	\begin{center}
		\includegraphics[width=\linewidth]{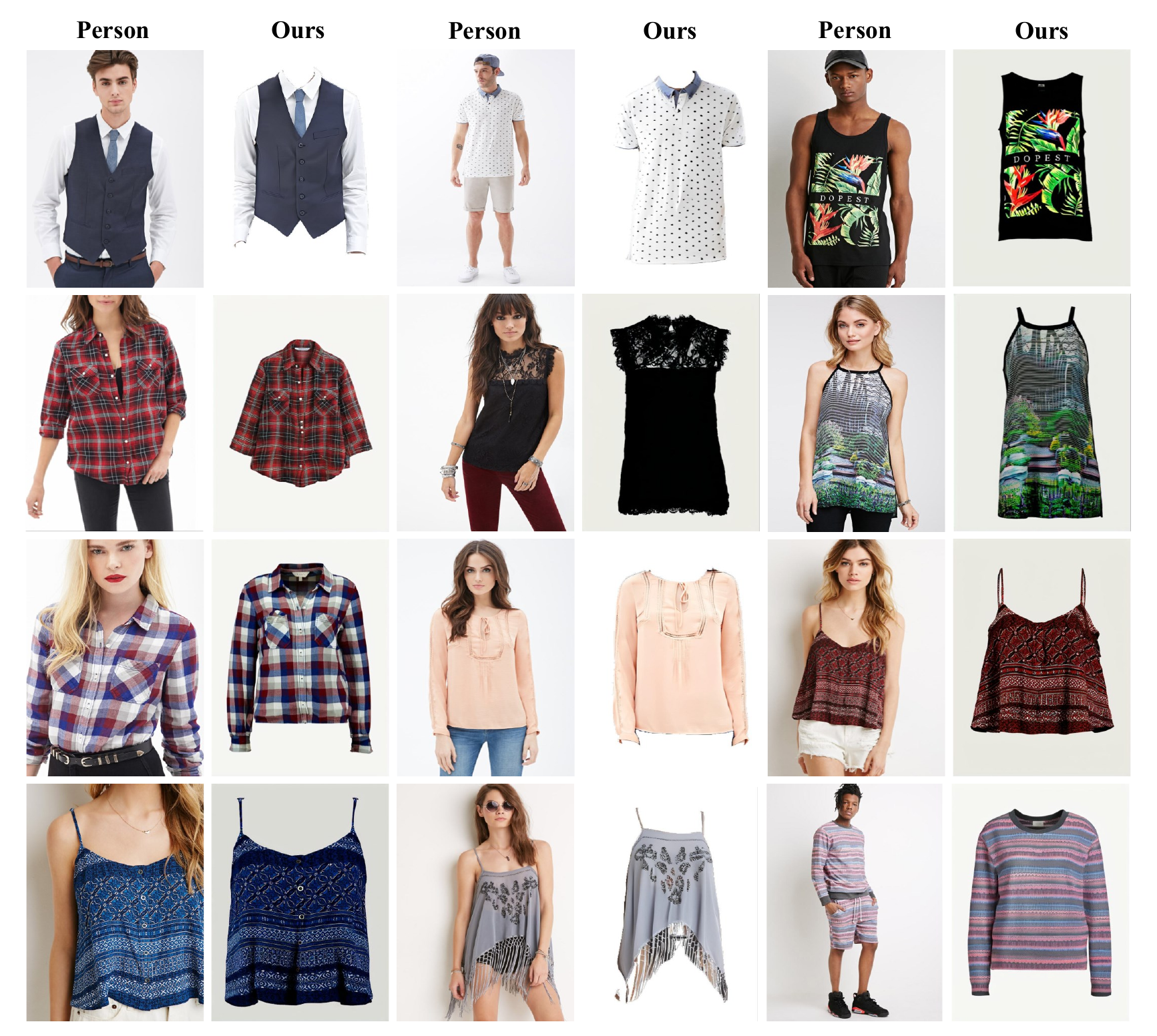} 
	\end{center}
    \caption{\textbf{Qualitative try-off results on DeepFashion-MultiModal.} \emph{Please zoom in to better observe the details.}}
 \label{fig:garment_deepfashion}
\end{figure*}

As shown in Fig.~\ref{fig:garment2}, we present additional try-off comparison results on the VITON-HD dataset. Additionally, we performed garment reconstruction on two open-source datasets DressCode and DeepFashion-MultiModal and visualized the quantitative results, as illustrated in Fig.~\ref{fig:garment_dresscode} and  Fig.~\ref{fig:garment_deepfashion}. As demonstrated above, our method effectively handles complex poses and occlusions, accurately restoring the garment's canonical shape while highly preserving its texture and structural details.

\subsection{User Study}
We conducted a user study with 50 participants using the model trained on the VITON-HD dataset. Each participant was randomly assigned 10 samples from a pool of 50 for evaluation, with each sample containing six different virtual try-on results generated in the person-to-person scenario.  Participants were asked to choose the best result using three criteria: image fidelity, human identity, and garment consistency. We totaled the number of times each method was chosen as the best across all test samples and calculated the average voting proportion for each method. As shown in Tab.~\ref{tab:user_study}, our method had the highest average voting proportion among all examples, indicating visually superior results and a significant advantage in human evaluation.
\setlength{\tabcolsep}{2pt}
\begin{table}[h]
    \small
    \centering
    \caption{\textbf{User study results.} We report the best-choice rate for our method and seven other methods, including StableVITON, OOTDiffusion, and CatVTON.}
    \begin{tabular*}{0.9\columnwidth}
    {@{\extracolsep{\fill}}l|cccc@{}}
    \hline
    \renewcommand{\arraystretch}{3.0}
     & OMFA & IDM-VTON & MV-VTON & others \\ \hline
     \textbf{Best  Choice Rate}& 46\%  & 8.6\% & 18.8\% & 26.6\% \\ 
    \hline
    \end{tabular*}
    \label{tab:user_study}
\end{table}

\section{Limitations}
Due to a lack of paired data for multi-layer garments, our proposed method does not provide multi-layer try-on/try-off. Furthermore, our architecture is only intended for a single garment input, whereas multiple garment inputs may dramatically extend the input sequence. In the future, we will incorporate more in-the-wild data to develop computationally efficient virtual try-on solutions that are more in line with real-world application scenarios.


\end{document}